\documentclass[
  journal=medium,
  manuscript=article-type,
  year=2020,
  volume=37,
]{cup-journal}

\usepackage{amsmath}
\usepackage{xcolor}
\usepackage[nopatch]{microtype}
\usepackage{booktabs}
\usepackage{algorithm} 
\usepackage{algpseudocode} 
\usepackage{amssymb}

\usepackage{cjhebrew}

\title{An Unsupervised Information-Theoretic Approach to Identifying Formulaic Clusters in Textual Data}

\author{Gideon Yoffe}
\affiliation{Department of Statistics and Data Science, The Hebrew University of Jerusalem}
\email[F. Author]{gideon.yoffe@mail.huji.ac.il}

\author{Yair Segev}
\affiliation{Faculty of Theology, Carl von Ossietzky Universität Oldenburg}

\author{Barak Sober}
\affiliation{Department of Statistics and Data Science, The Hebrew University of Jerusalem}


\begin{document}

\begin{abstract}
Texts, whether literary or historical, exhibit structural and stylistic patterns shaped by their purpose, authorship, and cultural context. Formulaic texts, which are characterized by repetition and constrained expression, tend to differ in their \textit{information content} (as defined by Shannon) compared to more dynamic compositions. Identifying such patterns in historical documents, particularly multi-author texts like the Hebrew Bible, provides insights into their origins, purpose, and transmission.
This study aims to identify formulaic clusters: sections exhibiting systematic repetition and structural constraints, by analyzing recurring phrases, syntactic structures, and stylistic markers. However, distinguishing formulaic from non-formulaic elements in an unsupervised manner poses a computational challenge, especially in high-dimensional, sample-poor data sets where patterns must be inferred without predefined labels.

To address this, we develop an information-theoretic algorithm that uses weighted \textit{self-information} distributions to recover structured partitions in text. The resulting clusters are interpreted from their self-information profiles and characteristic recurring features. By extending classical discrete self-information measures to a continuous formulation based on differential self-information in multivariate Gaussian distributions, our method remains applicable across various textual representations, including neural embeddings under Gaussian priors.

Applied to hypothesized authorial divisions in the Hebrew Bible, our approach isolates stylistic layers and provides a quantitative framework for textual stratification. This method enhances our ability to analyze compositional patterns, offering deeper insights into the literary and cultural evolution of texts shaped by complex authorship and editorial processes.
\end{abstract}

\section*{Plain Language Summary}

This work presents a method for identifying formulaic clusters—segments of text that rely heavily on repetition, structural regularity, or stylistic constraint. Such clusters can be found in a wide range of contexts, including religious texts, technical manuals, oral traditions, and institutional writing, where language is shaped by convention, genre, or the need for consistency. Detecting these patterns can offer insight into how texts were composed, transmitted, and shaped by cultural forces over time.

Our approach is grounded in information theory and focuses on the concept of \textit{self-information}, which measures how predictable or unpredictable an observation is under a given probability model. Recurring expressions are more probable, and therefore carry lower self-information, under the cluster-conditioned model in which they are common. The method separates text units according to differences in their cluster-conditioned feature distributions; the formulaic component is then interpreted from its self-information profile and concentration of recurring features.

Benchmarking against standard clustering methods on synthetic and real data demonstrates the method’s effectiveness, particularly under the sparse, high-dimensional configurations examined here. The framework uses separate discrete and Gaussian probability models appropriate to the corresponding data representations.

We apply our method to three books of the Hebrew Bible: Genesis, Exodus, and Leviticus. In Genesis, we focus on the distinction between genealogical lists and surrounding narrative material. The genealogical sections are known for their rigid, repetitive structure, and our method effectively isolates them as highly formulaic. In Exodus, we examine the division between Priestly (P) and non-Priestly texts, with the Priestly material comprising legal and cultic instructions. The method reliably separates these layers in line with traditional source-critical scholarship. In Leviticus, we test whether the method can distinguish between the Priestly source and the Holiness code (H), two stylistically distinct strata. The results show that subtle differences in recurring expressions and structural regularity allow for a clear separation between them under appropriate conditions.

This work offers a flexible, information-theoretic framework for analyzing the internal structure of texts. It is well-suited to the study of layered, edited, or composite documents, where patterns of repetition and constraint signal deeper organizational principles. The method is broadly applicable to literary, historical, legal, or technical corpora, wherever formulaic language plays a role in shaping textual form.

\section{Introduction}
\label{introduction}

An intriguing challenge in textual analysis is distinguishing formulaic layers—those characterized by recurring patterns, motifs, or stylistic conventions—from non-formulaic ones. In this context, a formulaic element is defined as a word or a sequence thereof, syntactic or grammatical structures, or thematic patterns that recur systematically within a corpus, either due to conventionalized linguistic usage, genre-specific constraints, or cultural transmission mechanisms. Formulaic expressions can function as structural scaffolds within texts, shaping their composition and reflecting underlying conventions of specific literary traditions, genres, or modes of authorship \citep[e.g.,][]{read2008measurement, paquot2012formulaic, wood2019classifying}. Beyond their structural significance, analyzing formulaic patterns provides insights into the dynamic interplay between creativity and constraint in language, illuminating the socio-cultural and cognitive forces that have influenced textual transmission and stylistic evolution \citep[e.g.,][]{magoun1953oral, jensen1980homeric}.

The importance of formulaicity is particularly pronounced in historical texts, where layers of composition often reflect complex authorship and redaction processes. Many historical documents, created and transmitted over centuries, are products of collective memory and institutional influence. As such, these texts often contain formulaic structures that encode cultural norms, narrative conventions, and theological or political ideologies \citep[e.g.,][]{knohl2007sanctuary, polak2017syntactic}. Understanding these layers can reveal how texts were constructed, transmitted, and adapted over time \citep[e.g.,][]{gitay1980tradition, polak2006linguistic, coleman2019psalmic}.

Among historical texts, the Hebrew Bible provides a compelling case study. As a composite text shaped by centuries of authorship, redaction, and transmission, the Hebrew Bible embodies a rich interplay of formulaic and non-formulaic elements. Its stylistic diversity reflects contributions from different schools of authorship, genres, and literary motifs, ranging from tightly structured legal codes to more fluid poetic and narrative traditions \citep{warters1976formula, shectman2009strata, stipp201711}. Detection and analysis of formulaic patterns within the Hebrew Bible offer a pathway to better understanding its compositional history, shedding light on how diverse voices and traditions coalesced into the canonical form.

Understanding the distinction between formulaic and non-formulaic layers requires rigorous methods for quantifying stylistic and structural patterns across literary feature sets. One promising metric for addressing these challenges is entropy, a concept from information theory that measures the unpredictability or randomness of a dataset. Initially introduced by Rudolf Clausius in 1865 as a measure of energy dispersal in a system, and subsequently formalized statistically by \citet{shannon1948mathematical}, entropy has found applications in diverse fields, including thermodynamics, cryptography, and data compression. In the context of textual analysis, entropy provides a framework for quantifying the variability and regularity inherent in linguistic structures. Formally, given a discrete probability distribution $P = \{p_1, p_2, \dots, p_n\}$ over $n$ possible elements, Shannon entropy is defined as:

\begin{equation}
H(P) = -\sum_{i=1}^{n} p_i \log p_i
,\end{equation}
where $p_i$ represents the probability of the $i$th element. Higher entropy values indicate greater variability, while lower entropy values correspond to constrained, repetitive structures, making entropy a powerful tool for distinguishing formulaic clusters from more unpredictable distributions.

The relevance of entropy to textual analysis lies in its ability to capture the degree of predictability of literary or linguistic features within a sequence, such as a sentence, paragraph, or chapter. Texts with high formulaicity, such as legal codes or liturgical compositions, are likely to exhibit low entropy, such that text units within them resemble one another, for example, in vocabulary or grammatical structure. Conversely, more creative or extemporaneous texts often exhibit higher entropy, reflecting their greater unpredictability. Information-based metrics, therefore, serve as a natural tool for distinguishing between formulaic and non-formulaic elements within texts \citep[e.g.,][]{church1990word}.

In textual contexts, entropy is closely related to the concept of perplexity---the exponential of the average self-information across all features in a sequence---which is widely used in language modeling and computational linguistics. Perplexity measures how well a probabilistic model predicts a sequence of words, with lower perplexity indicating a better fit. Formally, perplexity is defined as the exponential of the average negative log probability of the observed sequence:
\begin{equation} \label{eq_perplexity}
\text{Perplexity} = \exp\left(-\frac{1}{N} \sum_{i=1}^N \log P(w_i \mid w_{i-1}, w_{i-2}, \dots)\right),
\end{equation}
where $w_i$ represents the $i$th word in the sequence, and $P(w_i \mid w_{i-1}, w_{i-2}, \dots)$ denotes the conditional probability of $w_i$ given its preceding context.
Perplexity has proven instrumental in analyzing textual corpora over various linguistic angles \citep[e.g.,][]{klakow2002testing, huang2007dialect, gamallo2017perplexity}. In the context of formulaic analysis, perplexity can reveal how tightly a text adheres to specific linguistic conventions \citep[e.g.,][]{kurzynski2023stylometry}.

The task of identifying formulaic layers is inherently complex, as clustering formulaic elements within a corpus entails evaluating an exponentially large set of possible ways to divide the text, since each partition represents a different grouping of formulaic elements. This combinatorial challenge necessitates algorithmic approaches that efficiently approximate optimal solutions. One common strategy involves modeling linguistic feature distributions using probabilistic frameworks, with Gaussian-based methods being among the most widely applied.

For instance, Gaussian Mixture Models (GMMs) \citep{vlassis2002greedy} assume that the dataset can be represented as a weighted sum of Gaussian components, enabling efficient clustering through iterative optimization of parameters such as means, covariance matrices, and mixing weights. Similarly, Cross-Entropy Clustering (CEC) \citep{tabor2014cross} explicitly incorporates entropy as a criterion to partition data, refining clusters based on their statistical coherence. These methods, while effective in many contexts, rely on assumptions about the underlying data distribution, which may not always align with the structural and stylistic non-Gaussian variability inherent in textual corpora. 

Recent advances in representation learning, including the Contrastive Language–Image Pretraining model (CLIP; \citet{radford2021clip}) and Bidirectional Encoder Representations from Transformers (BERT; \citet{devlin2018bert}), have introduced powerful mechanisms for unsupervised clustering and alignment in high-dimensional semantic spaces. These models often optimize mutual information across paired or contextualized inputs, effectively capturing latent structural relationships without requiring explicit probabilistic assumptions. Building on these architectures, recent work has proposed information-theoretic evaluation metrics computed over embedding spaces to quantify diversity and structural complexity. One such class of methods includes entropy measures based on Rényi Kernel Entropy (RKE), which estimate entropy non-parametrically using similarity kernels to capture the dispersion of representations in high-dimensional spaces \citep{principe2010information, giraldo2014measures}. These approaches are increasingly used to assess representational diversity and clustering compactness in generative models and learned embeddings. Many of these entropy-based methods derive from or build upon neural embedding models such as variational autoencoders (VAEs), which assume that latent features follow a Gaussian distribution in a continuous space \citep{dilokthanakul2016deep, yang2019deep, li2020data}. VAEs excel at capturing the underlying structure of data by encoding inputs into compact latent representations governed by a Gaussian prior \citep{casale2018gaussian}, enabling smooth interpolation and the generation of new data points that resemble the original distribution.

While these approaches provide useful frameworks for clustering and representation learning, their reliance on Gaussian assumptions and on metrics that require estimating the covariance presents notable challenges. In textual data, where features often exhibit categorical, skewed, or multimodal distributions, the Gaussian prior may not accurately capture structural variability. Moreover, many of these methods are highly sensitive to the sample-to-dimension ratio \citep{altonji1996small, yao2015sample}. When the sample size is small relative to the feature space, covariance estimation becomes unstable, leading to poorly conditioned or singular matrices \citep[e.g.,][]{ashurbekova2021optimal}. Regularization techniques, such as adding small positive constants to the diagonal of covariance matrices, can partially mitigate these issues but introduce biases that require careful tuning \citep{bickel2008regularized}. This bias arises because regularization alters the estimated eigenvalues of the covariance matrix, effectively shrinking them toward a predetermined value, which can distort the data's true structure and influence clustering or classification outcomes.
Crucially, while these techniques are effective in high-resource, general-purpose NLP settings, their applicability to low-resource or singular historical corpora remains limited. In domains such as ancient languages, where pretrained embeddings are unreliable or unavailable, and where interpretability is essential for source-critical and literary-historical inquiry, alternative approaches are required.

These challenges are particularly pronounced in textual analysis, where statistical significance often necessitates dividing texts into relatively large units, such as paragraphs or entire chapters. While this improves the reliability of individual measurements, it also limits the number of available samples. At the same time, textual data often exhibit high dimensionality due to feature-rich representations, such as frequent words, $n$-grams, or syntactic constructions, leading to a severe imbalance between the number of samples and the number of features. This issue is especially prevalent in stylometric analysis, where subtle stylistic differences are captured through sparse, high-dimensional feature spaces \citep[][]{Stamatatos2009}. Given these constraints, clustering methods that rely on covariance estimation or strong parametric assumptions struggle to produce stable and interpretable results.

To address these limitations, we introduce an entropy-driven soft clustering framework that identifies formulaic structures based on the log-probability ($\log p$) of data points under a given distribution. The choice of $-\log p$ as our core metric stems from fundamental principles in information theory: the information content of an observation $x$, given a probability distribution $p(x)$, is defined as $-\log p(x)$, known as its \textit{self-information}. This quantity directly captures the degree of predictability of $x$—highly predictable elements (i.e., those frequently occurring in structured, formulaic segments) have higher probabilities and thus lower self-information values, whereas less structured, unpredictable elements exhibit lower probabilities and correspondingly higher self-information values.

Our method evaluates observations under cluster-conditioned probability models, allowing candidate partitions to be characterized through differences in feature distributions and sample-wise self-information. The framework uses soft sample weights to represent candidate cluster memberships and provides separate objectives for discrete and continuous representations.
The discrete formulation estimates feature probabilities directly and does not require covariance estimation. The continuous Gaussian formulation, by contrast, estimates and inverts a regularized weighted covariance matrix. The two formulations, therefore, address different data types through different probabilistic models. Any numerical advantage of the continuous formulation must be established empirically and does not arise from avoiding covariance estimation or inversion.

To accommodate different types of textual feature representations, we consider two complementary formulations of the method: one for discrete data and one for continuous data. The discrete formulation operates directly on binary or count-based features, avoiding any form of covariance estimation. In contrast, the continuous formulation, used for real-valued vector representations, such as embeddings, does involve estimating a soft covariance matrix. While this introduces some of the challenges seen in classical approaches, our method confines the estimation to weighted subsets of the data associated with candidate clusters exhibiting internal coherence. The continuous formulation estimates a regularized weighted covariance matrix for each candidate cluster and therefore remains sensitive to small effective sample sizes. In both formulations, the objective is expressed in terms of cluster-conditioned log probabilities, using a probability model appropriate to the corresponding data representation. This provides a common information-theoretic framework for discrete and continuous data while retaining their distinct distributional assumptions.

The paper is structured as follows: In \S\ref{algorithm}, we introduce the mathematical foundations of the information-based soft clustering framework, defining the self-information distribution and formulating an optimization scheme for distinguishing formulaic from non-formulaic textual components in \S\ref{benchmarking}, we benchmark the method against existing clustering techniques on synthetic datasets, assessing its ability to classify structured and unstructured elements under controlled conditions, in \S\ref{results} we apply the framework to the analysis of the biblical corpus, evaluating its capacity to differentiate hypothesized literary strata based on stylistic and formulaic properties, and in \S\ref{conclusion} we conclude.

\section{Soft Clustering Algorithm}
\label{algorithm}

The following sections establish the mathematical foundations of the proposed information-based soft clustering framework and formulate an optimization scheme for distinguishing formulaic from non-formulaic text layers. In this framework, "soft clustering" refers to the relaxation techniques employed during optimization to avoid indeterminate solutions \citep[e.g.,][]{royer2017adaptive}. This approach follows established clustering methodologies, in which constraints are relaxed to transform discrete, hard-assignment problems into continuous optimization problems, making them more tractable. An example of this is semidefinite relaxation in graph clustering, which reformulates discrete clustering problems as continuous ones, enabling more efficient optimization. By utilizing such relaxation techniques, our framework can more effectively identify structured patterns in text while maintaining flexibility in handling ambiguity in cluster assignments.

\subsection{Self-Information as a Measure for Identifying Formulaic Structures}

The use of self-information stems from its foundational role in information theory, where the information content of an observation $x$ is defined as:
\begin{equation}
    I(x) = -\log p(x),
\end{equation}
which quantifies how predictable or surprising an observation is. More predictable elements (those that occur frequently in structured, formulaic segments) have higher probabilities and thus lower $-\log p(x)$ values, while less predictable elements exhibit lower probabilities and higher $-\log p(x)$ values.
In textual analysis, a recurring feature has lower self-information in the cluster in which it is common. This does not imply that every segment in that cluster has a lower mean self-information, because the segment-level value includes contributions from all of its features, including rare ones. The discrete objective partitions samples according to their likelihood under the two cluster-conditioned feature distributions; it does not rank the clusters by their mean self-information. We identify the formulaic cluster from the recurrence and concentration of its distinctive $n$-grams.

\paragraph{\textbf{Layman explanation:}} Frequently recurring expressions are less surprising and therefore have lower self-information where they are common. The algorithm separates text units according to differences in their feature distributions. We then examine the recurring features of each cluster to determine which cluster exhibits a formulaic structure.

\vspace{1em}

\subsection{Clustering Approach}
For a two-cluster problem, clustering methods aim to partition a dataset $X = \{x_1, x_2, \dots, x_n\}$ into two disjoint subsets, or clusters, based on a given similarity measure. Traditional hard clustering assigns each data point $x_i$ to exactly one of the two clusters $\mathcal{C}_1$ or $\mathcal{C}_2$ such that:
\begin{equation}
    x_i \in \mathcal{C}_j \quad \text{where} \quad \mathcal{C}_1 \cup \mathcal{C}_2 = X, \quad \mathcal{C}_1 \cap \mathcal{C}_2 = \emptyset.
\end{equation}

Soft clustering relaxes this constraint by allowing each data point $x_i$ to have a continuous membership weight $s_i \in [0,1]$, representing its degree of association with one of the two clusters. Instead of assigning $x_i$ to a single cluster, we define:
\begin{equation}
    s_i \in [0,1], \quad s_i = 0 \text{ for full membership in } \mathcal{C}_1, \quad s_i = 1 \text{ for full membership in } \mathcal{C}_2.
\end{equation}
Thus, each data point has a soft assignment, where
\[
\begin{cases} 
x_i \in \mathcal{C}_1, & \text{if } s_i = 0, \\
x_i \in \mathcal{C}_2, & \text{if } s_i = 1, \\
x_i \in \mathcal{C}_1 \text{ and } \mathcal{C}_2 \text{ with weight } (1 - s_i, s_i), & \text{if } 0 < s_i < 1.
\end{cases}
\]

Our clustering approach employs a relaxation strategy within the optimization process, allowing for continuous rather than binary sample assignments. At each optimization step, we update the membership of each sample in one of two candidate clusters. Because the cluster labels are exchangeable, neither candidate is designated as formulaic during optimization. After convergence, the continuous memberships are thresholded once to obtain a binary partition, and the resulting clusters are characterized through their cluster-conditioned self-information and distinctive features.

The role of self-information differs slightly between the two formulations. The continuous objective explicitly favors a sufficiently populated candidate cluster with low and narrowly distributed self-information. The discrete objective instead separates samples according to their likelihood under two cluster-conditioned feature distributions. In the discrete application, the recovered partition is subsequently oriented using self-information diagnostics, while its interpretation as formulaic is supported by the concentration of recurring features.

\paragraph{\textbf{Layman explanation:}} Rather than forcing each text unit immediately into one cluster or the other, the method allows intermediate degrees of membership while searching for a partition. These weights do not identify a formulaic cluster in advance. Once the partition has converged, the memberships are converted into two clusters, whose self-information profiles and recurring features are then examined to determine which exhibits formulaic structure.

\vspace{1em}

\subsection{Clustering Formalism for Discrete Categorical Data} \label{formalism_onehot}

To determine optimal cluster assignments, we minimize the cross-entropy between the empirical data distribution $\mathcal{P}_{\text{data}}(x)$ and the model distribution $\mathcal{P}_{\text{model}}(x)$. The empirical distribution $\mathcal{P}_{\text{data}}(x)$ represents the observed frequencies of different textual elements in the dataset, while $\mathcal{P}_{\text{model}}(x)$ is derived from the soft clustering assignments, representing the likelihood of each sample belonging to a given cluster. Minimizing cross-entropy ensures that $\mathcal{P}_{\text{model}}(x)$ closely approximates the observed data, allowing the model to generalize the underlying structure while maintaining probabilistic flexibility.

\paragraph{\textbf{Layman explanation:}} The goal is to ensure that the feature distributions within each cluster align well with the actual data. Cross-entropy serves as a standard measure of divergence between observed and modeled probabilities. Minimizing it improves the model’s ability to describe how features are distributed in the data.

\vspace{1em}

Formally, cross-entropy is defined as:

\begin{equation} \label{eq:cross_entropy_expectation}
    H(\mathcal{P}_{\text{data}}, \mathcal{P}_{\text{model}}) = - \mathbb{E}_{\mathcal{P}_{\text{data}}} \left[ \log \mathcal{P}_{\text{model}}(x) \right] = - \sum_{x} \mathcal{P}_{\text{data}}(x) \log \mathcal{P}_{\text{model}}(x).
\end{equation}

For discrete data, we use probability models appropriate to the feature representation. Binary presence--absence features are modeled using Bernoulli distributions, feature-specific counts from known numbers of trials using Binomial distributions, and count vectors conditional on their totals using Multinomial distributions. In each case, the cluster-conditioned feature probabilities are estimated from frequencies weighted by the soft sample assignments. The Bernoulli formulation is used for the synthetic categorical benchmark, while the Multinomial formulation is used for the biblical analyses; the Binomial formulation is included as an additional extension.

\paragraph{\textbf{Layman explanation:}} The Bernoulli model records whether each feature is present, the Binomial model records its occurrences among a known number of opportunities, and the Multinomial model records how a sample's total count is distributed among the available features. We use the latter when analyzing cumulative $n$-gram counts in the biblical texts.

\vspace{1em}

Under the Bernoulli formulation, the soft probability that the $j$th feature is present in a given cluster is estimated as

\begin{equation} \label{eq:soft_freq_discrete}
    p_{\mathrm{B}}(w_j | \vec{s})
    =
    \frac{\sum_{i=1}^{n} s_i x_{ij}}
         {\sum_{i=1}^{n} s_i},
\end{equation}
where $x_{ij}=1$ if the $j$th feature is present in sample $x_i$ and $x_{ij}=0$ otherwise. The probability for the complementary cluster is obtained by replacing $\vec{s}$ with $1-\vec{s}$. Unlike a categorical probability vector, Bernoulli probabilities are not normalized across features.

\paragraph{\textbf{Layman explanation:}} This probability records how frequently each feature is present among samples assigned to a cluster. Because several features may occur in the same sample, their probabilities need not sum to one.

\vspace{1em}

With these estimated feature probabilities, we can compute the likelihood of a complete sample given a cluster. Under the Bernoulli assumption, the probability of a sample $x_i$ conditioned on its cluster assignment is:

\begin{equation} \label{eq:bernoulli_prob_formulaic}
    P_{\mathrm{B}}(x_i | \vec{s}) = \prod_{j=1}^{m} p_{\mathrm{B}}(w_j | \vec{s})^{x_{ij}} \left[1 - p_{\mathrm{B}}(w_j | \vec{s})\right]^{1 - x_{ij}},
\end{equation}

\begin{equation} \label{eq:bernoulli_prob_non_formulaic}
    P_{\mathrm{B}}(x_i | 1 - \vec{s}) = \prod_{j=1}^{m} p_{\mathrm{B}}(w_j | 1 - \vec{s})^{x_{ij}} \left[1 - p_{\mathrm{B}}(w_j | 1 - \vec{s})\right]^{1 - x_{ij}}.
\end{equation}

Here, $p_{\mathrm{B}}(w_j | \vec{s})$ and $p_{\mathrm{B}}(w_j | 1 - \vec{s})$ are the \textit{soft} Bernoulli probabilities of the $j$th feature being present in each cluster.

\paragraph{\textbf{Layman explanation:}} These expressions compute how likely a full set of features is for a given sample under each cluster model. If a text segment contains many features that are common in one cluster but rare in the other, the value will reflect that asymmetry.

\vspace{1em}

For any of the probability models considered below, the score is the negative cluster-weighted log-likelihood:

\begin{equation} \label{eq:score_function}
    S(\vec{s}, X) = -\sum_{i=1}^{n} \left( s_i \cdot \log P(x_i | \vec{s}) + (1 - s_i) \cdot \log P(x_i | 1 - \vec{s}) \right),
\end{equation}

where $\vec{s} \in [0,1]^n$ is the weight vector representing the soft assignment of samples to clusters. For the Bernoulli model, $X\in\{0,1\}^{n\times m}$; for the Binomial and Multinomial models, $X\in\mathbb{N}_0^{n\times m}$.

\paragraph{\textbf{Layman explanation:}} The score function evaluates how well the current cluster assignments explain the data. The algorithm updates these assignments to minimize the score, thereby improving the model’s alignment with the observed feature patterns.

\vspace{1em}

To accommodate non-binary data, we distinguish between two count models. The Binomial model describes feature-specific counts from a known number of trials, whereas the Multinomial model describes a count vector conditional on its total count. Together with the Bernoulli formulation above, these models cover binary, trial-based, and cumulative-count representations. For compactness, let $\vec{r}\in\{\vec{s},1-\vec{s}\}$ denote either candidate cluster.

\paragraph{Binomial formulation.}

The Binomial model applies when $x_{ij}$ is the number of occurrences of feature $j$ among $N_{ij}$ known trials, with $0\leq x_{ij}\leq N_{ij}$. Its cluster-conditioned probability is
\begin{equation}
\label{eq:soft_freq_binomial}
p_{\mathrm{Bin}}(w_j|\vec{r})
=
\frac{\sum_{i=1}^{n}r_i x_{ij}}
     {\sum_{i=1}^{n}r_i N_{ij}}.
\end{equation}
The corresponding sample likelihood is
\begin{equation}
\label{eq:binomial_prob}
P_{\mathrm{Bin}}(x_i|\vec{r})
=
\prod_{j=1}^{m}
\binom{N_{ij}}{x_{ij}}
p_{\mathrm{Bin}}(w_j|\vec{r})^{x_{ij}}
\left[1-p_{\mathrm{Bin}}(w_j|\vec{r})\right]^{N_{ij}-x_{ij}},
\end{equation}
giving
\begin{align}
\label{eq:log_likelihood_binomial}
\log P_{\mathrm{Bin}}(X|\vec{r})
&=
\sum_{i=1}^{n}\sum_{j=1}^{m}
\Bigg[
\log\binom{N_{ij}}{x_{ij}}
+x_{ij}\log p_{\mathrm{Bin}}(w_j|\vec{r})
\nonumber\\
&\qquad
+(N_{ij}-x_{ij})
\log\left[1-p_{\mathrm{Bin}}(w_j|\vec{r})\right]
\Bigg].
\end{align}

\paragraph{Multinomial formulation.}

For a count vector, let
\begin{equation}
n_i=\sum_{j=1}^{m}x_{ij}
\end{equation}
denote the total number of feature occurrences in sample $x_i$. Conditional on $n_i$, the cluster-conditioned feature probabilities are
\begin{equation}
\label{eq:soft_freq_multinomial}
p_{\mathrm{M}}(w_j|\vec{r})
=
\frac{\sum_{i=1}^{n}r_i x_{ij}}
     {\sum_{i=1}^{n}r_i n_i},
\qquad
\sum_{j=1}^{m}p_{\mathrm{M}}(w_j|\vec{r})=1.
\end{equation}
The corresponding sample likelihood is
\begin{equation}
\label{eq:multinomial_prob}
P_{\mathrm{M}}(x_i|\vec{r})
=
\frac{n_i!}{\prod_{j=1}^{m}x_{ij}!}
\prod_{j=1}^{m}
p_{\mathrm{M}}(w_j|\vec{r})^{x_{ij}},
\end{equation}
giving
\begin{equation}
\label{eq:log_likelihood_multinomial}
\log P_{\mathrm{M}}(X|\vec{r})
=
\sum_{i=1}^{n}
\left[
\log(n_i!)
-\sum_{j=1}^{m}\log(x_{ij}!)
+\sum_{j=1}^{m}x_{ij}\log p_{\mathrm{M}}(w_j|\vec{r})
\right].
\end{equation}

Substituting the appropriate cluster-conditioned likelihood into Eq.~\eqref{eq:score_function} gives the Bernoulli, Binomial, or Multinomial objective. The Multinomial formulation is used for the cumulative $n$-gram count matrices in the biblical analyses.

\paragraph{\textbf{Layman explanation:}}
The Bernoulli model records whether a feature occurs, the Binomial model records how often it occurs among a known number of opportunities, and the Multinomial model records how a sample's total feature count is distributed among the available features.


\vspace{1em}

While this formulation provides a principled probabilistic basis for clustering, it does not explicitly account for competing structural signals that may emerge from different generative processes within the data. However, assessing the presence and dominance of such alternative signals is beyond the scope of this work, as our primary focus is on the probabilistic clustering framework itself. That said, the relative influence of entropy-based clustering versus clustering driven by feature distributions depends on dataset-specific parameters, such as the baseline activation probability of features and the presence of structured, formulaic dimensions. A heuristic analysis of this interplay is provided in \ref{app_feat_norm}, where we illustrate how these factors affect clustering outcomes and discuss the need for future work to develop a predictive framework for determining when each signal will dominate.

\subsection{Clustering Formalism for Continuous Multivariate Gaussian Data}

In the case of continuous data, applying a Bernoulli model is not feasible, as such data does not inherently exhibit discrete binary properties. Instead, we adopt an information-based formalization, which provides a natural and mathematically elegant means of characterizing the structure of high-dimensional continuous distributions. Specifically, for data modeled as multivariate Gaussians, entropy offers a well-parameterized measure of dispersion and structure, making it a suitable alternative for defining soft cluster assignments.

\paragraph{\textbf{Layman explanation:}} In the discrete case, each text segment is represented by binary indicators, capturing the presence or absence of individual features (e.g., words or $n$-grams) in a text unit. However, this representation does not extend to settings where features are continuous-valued, such as neural embeddings or normalized frequency vectors. In these cases, one must model how the values themselves vary across samples rather than simply whether a feature occurs. The multivariate Gaussian distribution provides a natural and mathematically grounded way to describe such continuous variation. Within this framework, entropy reflects how concentrated or dispersed the distribution is, offering a principled measure of structural regularity in continuous feature space.

\vspace{1em}

Formally, in a $d$-dimensional space, the entropy of a multivariate Gaussian distribution is given by
\begin{equation}
    H(X) = \frac{d}{2}(1 + \ln(2\pi)) + \frac{1}{2} \ln |\Sigma|,
\end{equation}
where $\Sigma$ is the covariance matrix, and $|\Sigma|$ denotes its determinant. 

To incorporate weights in our soft clustering framework, we define a weighted covariance matrix:
\begin{equation}
    \mu_w = \frac{\sum_{i=1}^n s_i x_i}{\sum_{i=1}^n s_i}, \quad
    \Sigma_w = \frac{\sum_{i=1}^n s_i (x_i - \mu_w)(x_i - \mu_w)^\top}{\sum_{i=1}^n s_i},
\end{equation}
where $\{s_i\}$ are the weights assigned to samples. To enhance numerical stability in cases where the covariance matrix may be ill-conditioned, a regularization term is applied:
\begin{equation}
    \Sigma_w \gets \Sigma_w + \epsilon I,
\end{equation}
where $\epsilon$ is a small positive constant, and $I$ is the identity matrix of rank $d$. Using this regularized covariance, the total entropy of the weighted dataset is computed as:

\begin{equation}
    H(X) = \frac{d}{2}(1 + \ln(2\pi)) + \frac{1}{2} \ln |\Sigma_w|.
\end{equation}

This expression provides a global measure of the dataset's uncertainty based on its covariance structure. However, in the context of soft clustering, we require a measure that quantifies the contribution of individual samples to the overall uncertainty. Since entropy quantifies unpredictability, a natural way to approximate the uncertainty associated with a specific sample $x_{i'}$ is to consider its likelihood under the dataset’s distribution.

For a multivariate Gaussian, the log-likelihood of a sample is:

\begin{equation}
    \log P(x_{i'}|\vec{s}) = -\frac{d}{2} \log(2\pi) - \frac{1}{2} \log |\Sigma_w| - \frac{1}{2} (x_{i'} - \mu_w)^\top \Sigma_w^{-1} (x_{i'} - \mu_w).
\end{equation}

This decomposition separates the likelihood into three terms: (i) a normalization constant, (ii) a term encoding the dataset’s covariance structure, and (iii) a term measuring a sample’s deviation from the mean under the Mahalanobis metric. Subsequently, the self-information of a sample is

\begin{equation}
    I(x_{i'}|\vec{s}) = -\log p(x_{i'}) = \frac{d}{2} \log(2\pi) + \frac{1}{2} \log |\Sigma_w| + \frac{1}{2} D_M(x_{i'}|\vec{s}),
\end{equation}

where $D_M(x_{i'}|\vec{s}) = (x_{i'} - \mu_w)^\top \Sigma_w^{-1} (x_{i'} - \mu_w)$ is the Mahalanobis distance given the weighting by $\vec{s}$ (see \ref{app_formalism_gaussian}). 

\paragraph{\textbf{Layman explanation:}} A sample’s self-information increases the further it lies from the mean of the data, especially if it does so along directions where the data varies less. This distance, measured by the Mahalanobis metric, provides a principled way of quantifying how typical or atypical a given point is relative to the distribution of other points.

\vspace{1em}

\noindent
This equation directly relates a sample’s entropy contribution to its likelihood under the dataset’s distribution.

The Mahalanobis distance quantifies how far a sample deviates from the mean, making it a natural proxy for self-information. Samples close to $\mu_w$ have higher likelihoods and lower entropy, while those further away contribute greater uncertainty. This formulation provides a probabilistic approach to clustering, in which samples are distinguished by their deviation from the cluster-conditioned covariance structure. 

\paragraph{Score Function for Cluster Identification}

To recover candidate partitions from Gaussian data, we define the heuristic score
\begin{equation}\label{eq_score_gaussian}
    S(I(X\mid\vec{s}))
    =
    \operatorname{std}(I(X\mid\vec{s}))
    \cdot
    \operatorname{avg}(I(X\mid\vec{s}))
    \cdot
    \frac{1}{\lVert\vec{s}\rVert_2},
\end{equation}
where $I(X\mid\vec{s})=\{I(x_i\mid\vec{s})\}_{i=1}^{n}$ contains the self-information values of all samples under the weighted Gaussian model defined above. The operators $\operatorname{std}$ and $\operatorname{avg}$ denote their empirical standard deviation and mean, respectively.

\paragraph{\textbf{Layman explanation:}} The score function favors clusters that are internally consistent (low variability), strongly predictable (low average self-information), and include enough samples to be meaningful. These three criteria together help isolate regions of the data that exhibit formulaic regularity.

\vspace{1em}

This score combines the mean and standard deviation of the self-information distribution with the inverse norm of the sample weights. The mean quantifies the typical predictability of the selected samples, the standard deviation quantifies their internal consistency, and the inverse-norm term discourages trivially small clusters. Minimizing the score, therefore, favors a sufficiently populated candidate cluster with low and narrowly distributed self-information, as expected for formulaic structure.

Two aspects of this formulation are noteworthy. (\textbf{1}) Gaussian cross-entropy clustering \citep[e.g.,][]{tabor2014cross} requires estimating the covariance matrix, which can be unstable when the sample size is small relative to the dimensionality of the feature space. Our continuous formulation also estimates and inverts a weighted covariance matrix, applying regularization to ensure invertibility. Regularization does not eliminate finite-sample estimation error. The distinction instead lies in the objective: our score evaluates the mean and dispersion of sample-wise self-information within a candidate cluster while penalizing solutions with insufficient membership. The benchmarking results show that this objective benefits empirically from increasing sample size and dimensionality under the covariance-scale configurations considered here.
(\textbf{2}) Self-information has the same interpretation in the Gaussian and discrete formulations. Because $I(x)=-\log p(x)$, observations that are more probable under their cluster-conditioned model have lower self-information. A concentrated and predictable Gaussian cluster is therefore expected to exhibit low and narrowly distributed self-information. Likewise, recurring discrete features have lower self-information under the cluster in which they are common. Sparsity may cause individual rare activations to have high self-information, but it does not reverse the relationship between probability and self-information. The discrete objective described in \S\ref{formalism_onehot} identifies differences between cluster-conditioned feature distributions; the interpretation of a recovered cluster as formulaic is additionally supported by its recurring features and self-information profile.


\subsection{Optimization Scheme} \label{optim}

The algorithm optimizes weights $\vec{s}\in[0,1]^n$ to recover a candidate partition from the cluster-conditioned likelihoods. In the discrete formulation, the two cluster labels are exchangeable, and formulaicity is inferred from the recovered self-information profiles and recurring features after convergence. The optimization scheme is summarized in Algorithm~\ref{alg_optim}. While our method is flexible and conceptually transparent, it is also more computationally intensive than conventional iterative clustering algorithms such as $k$-means or Expectation-Maximization (EM). Those methods benefit from closed-form updates for centroids or covariance matrices, leading to fast convergence in low-dimensional, well-behaved data. In contrast, our approach involves nonlinear optimization over a continuous weight vector. This added cost reflects the complexity of modeling self-information and structural predictability, rather than geometric proximity or global variance. In scenarios where sample sizes are large and feature dimensionality is low, iterative methods may indeed be more efficient and comparably effective. However, this is rarely the case in historical or literary corpora, which tend to exhibit sparse, high-dimensional feature spaces and relatively small sample counts. It is precisely these conditions that motivated the design of an information-theoretic clustering approach tailored to such data.

\begin{algorithm}
\caption{\textbf{Candidate Partition Optimization Scheme}} 
	\begin{algorithmic}[1]
            \State Initialize $\vec{s}$ randomly, with $s_i\in(0,1)$ for all $i$.
            \While {not converged}
            \State Compute the model parameters and score associated with the selected formulation.
            \State Compute the corresponding score.
            \State Update $\vec{s}$ using a constrained numerical method, with $s_i\in[0,1]$.
            \EndWhile
            \State Set $z_i=1$ if $s_i\geq 1/2$ and $z_i=0$ otherwise.
            \State Characterize the recovered clusters from their self-information profiles and distinctive features.
            \State \textbf{return} $\vec{z}$.
	\end{algorithmic} 
\label{alg_optim}
\end{algorithm}

\section{Benchmarking} \label{benchmarking}

\subsection{Clustering Benchmarking on Categorical Data} \label{experiment_onehot}

To evaluate the resolving power of our algorithm on one-hot-encoded data, we design an experiment that generates synthetic datasets with features that reflect textual embeddings. These datasets simulate challenges such as sparsity, feature dependencies, and formulaic activation patterns.

The dataset is generated as follows: For each cluster, we create $n$ samples with dimensionality $d$, where features are binary, representing the activation (1) or absence (0) of specific elements. One cluster, referred to as the uniform cluster, has an equally small activation probability $p$ across all dimensions, resulting in sparse, randomly distributed features. The other cluster, referred to as the formulaic cluster, increases the activation probability for a subset of $d_{\rm{form}}$ dimensions by a bias factor $f$, while the remaining dimensions maintain the uniform activation probability, such that $p_{\text{form}} = p + f$. 

To introduce feature interdependencies within the formulaic cluster, we enforce correlations between pairs of dimensions. Specifically, for each sample in the formulaic cluster, $m$ dimensions are randomly selected from the formulaic subset, and the activation state of one dimension is set to match the other. This process ensures that certain features are not independent but instead exhibit a structured co-occurrence pattern. Such interdependencies mimic the stylistic or structural dependencies commonly found in textual data, where the presence of one linguistic feature often implies the presence of another.

However, it is important to acknowledge the inherent difficulty of modeling textual data. Texts are complex, with nuanced structures and dependencies that cannot be fully captured by simplistic binary features or synthetic generation processes. Despite these limitations, the purpose of this benchmark is not to replicate the textual data exactly but to demonstrate our algorithm's ability to solve a difficult nonlinear problem—optimizing the distinction between sparse, discrete datasets that differ in entropy. This serves as proof of the algorithm's robustness in handling challenging clustering problems.

\begin{figure}[t!]
\centering
{\includegraphics[scale = 0.45]{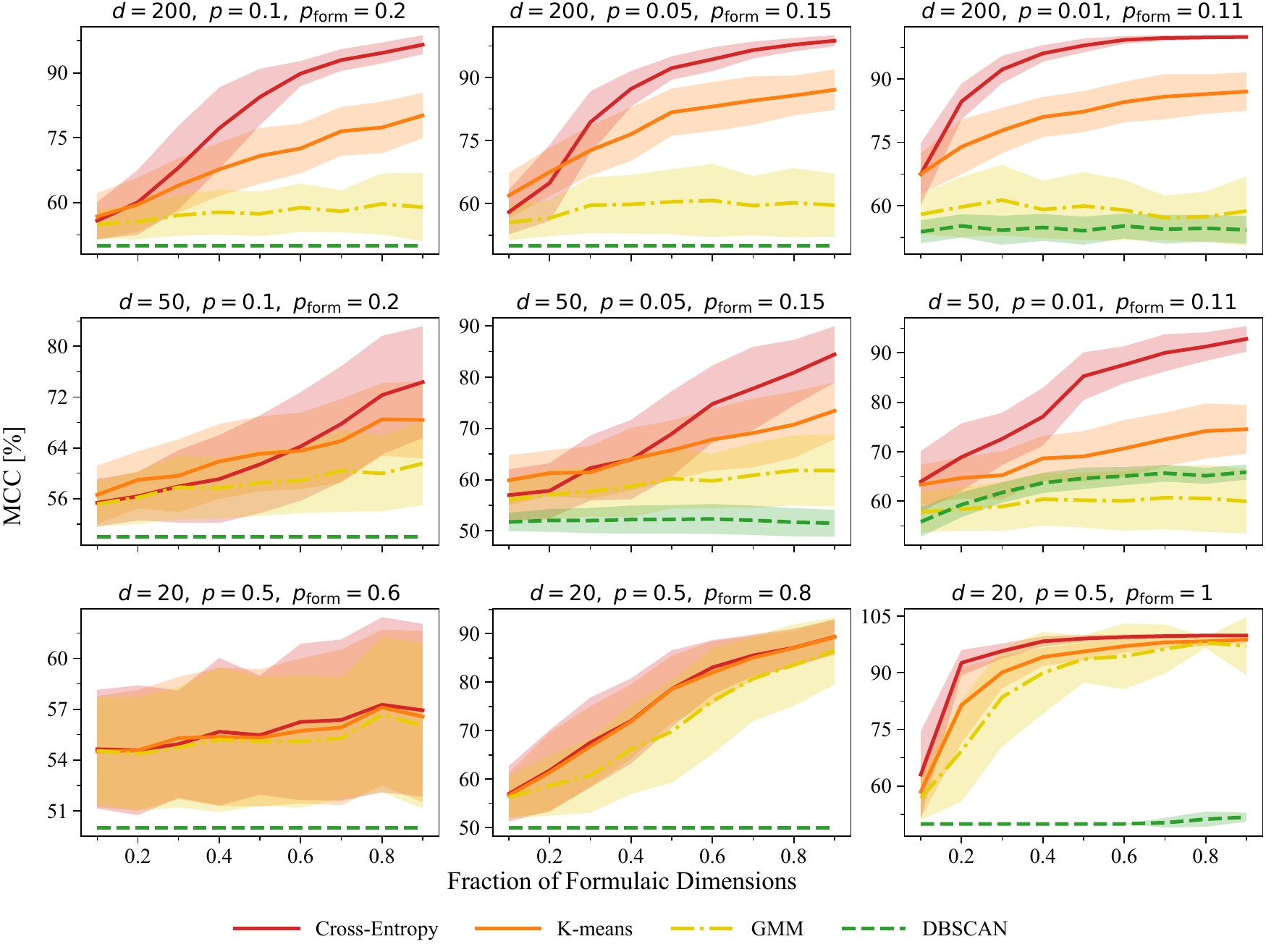}}
  \caption{Classification results for the benchmarking experiment of discrete categorical one-hot encoded discrete data described in \S\ref{experiment_onehot}. The test datasets included 100 samples of (equally-sized) formulaic and non-formulaic classes, of 200 (\textbf{top panel}), 50 (\textbf{middle panel}), and 20 dimensions (\textbf{bottom panel}), with varying degrees of the probability of the base- ($p$) and formulaic- feature activation ($p_{\text{form}}$), respectively, and the fraction of formulaic dimensions in the formulaic class. The colored areas represent one-standard-deviation intervals derived from 100 simulations.}
     \label{fig_mcc_onehot}
\end{figure}

We consider three nominal clustering algorithms that represent different clustering paradigms: (\textbf{1}) GMM, which assumes data is generated from a mixture of Gaussian distributions, making it suitable for probabilistic clustering in continuous spaces, (\textbf{2}) $k$-means \citep{hartigan1979k}, which partitions data into clusters based on minimizing intra-cluster variance, making it effective for Euclidean distance-based separation, and (\textbf{3}) DBSCAN \citep{ester1996density}, which identifies clusters as dense regions of points separated by low-density areas, making it robust to arbitrary-shaped clusters but sensitive to parameter tuning. The results of this experiment, shown in Figure~\ref{fig_mcc_onehot}, demonstrate the superior performance of our algorithm in resolving formulaic clusters in sparse categorical encoded datasets whose sample-to-dimension ratio is small.


The classification accuracy is measured using the Matthews Correlation Coefficient (MCC), given by
\begin{equation}
\label{eq_MCC}
\mathrm{MCC}
=
\frac{
\mathrm{TP}\,\mathrm{TN}-\mathrm{FP}\,\mathrm{FN}
}{
\sqrt{
(\mathrm{TP}+\mathrm{FP})
(\mathrm{TP}+\mathrm{FN})
(\mathrm{TN}+\mathrm{FP})
(\mathrm{TN}+\mathrm{FN})
}
}.
\end{equation}
Here, $\mathrm{TP}$, $\mathrm{TN}$, $\mathrm{FP}$, and $\mathrm{FN}$ denote the entries of the binary confusion matrix. Because cluster labels are exchangeable, we evaluate agreement using the absolute MCC and report
\begin{equation}
    \mathrm{MCC}_{\mathrm{norm}}
    = 50\left(1+\left|\mathrm{MCC}\right|\right)\%.
\end{equation}
Thus, $50\%$ indicates no association, whereas $100\%$ indicates perfect agreement up to an exchange of the cluster labels.
For concision, this normalized label-invariant quantity is referred to as MCC in the figures and results below.
Under the sparse, high-dimensional configurations tested, our method often outperforms the comparison algorithms in recovering the planted formulaic structure, although this advantage is not uniform across the full parameter range (Figure~\ref{fig_mcc_onehot}).

\subsection{Clustering Benchmarking on Multivariate Gaussian Data} \label{experiment_gaussian}

We perform a series of classification experiments to demonstrate the classification power of our algorithm compared with well-established routines for multivariate Gaussian data. These methods are: \textbf{(1)} Differential-Entropy \citep{davis2006differential}, \textbf{(2)} Gaussian Mixture Expectation-Maximization\footnote{\url{https://scikit-learn.org/1.5/modules/mixture.html}} (GMM), \textbf{(3)} Regularized Expectation-Maximization (Reg. EM) \citep{houdouin2023regularized}, and \textbf{(4)} Cross-Entropy-Clustering (CEC) \citep{tabor2014cross}. Additionally, we test the resolving power of the $L_2$ norm distribution by incorporating it into an optimization scheme similar to that described for the self-information distribution, given its relevance in high-dimensional settings, where the concentration of measure ensures that the norm becomes a distinguishing feature (but is independent of the number of samples).

\begin{figure}[t!]
\centering
{\includegraphics[scale = 0.45]{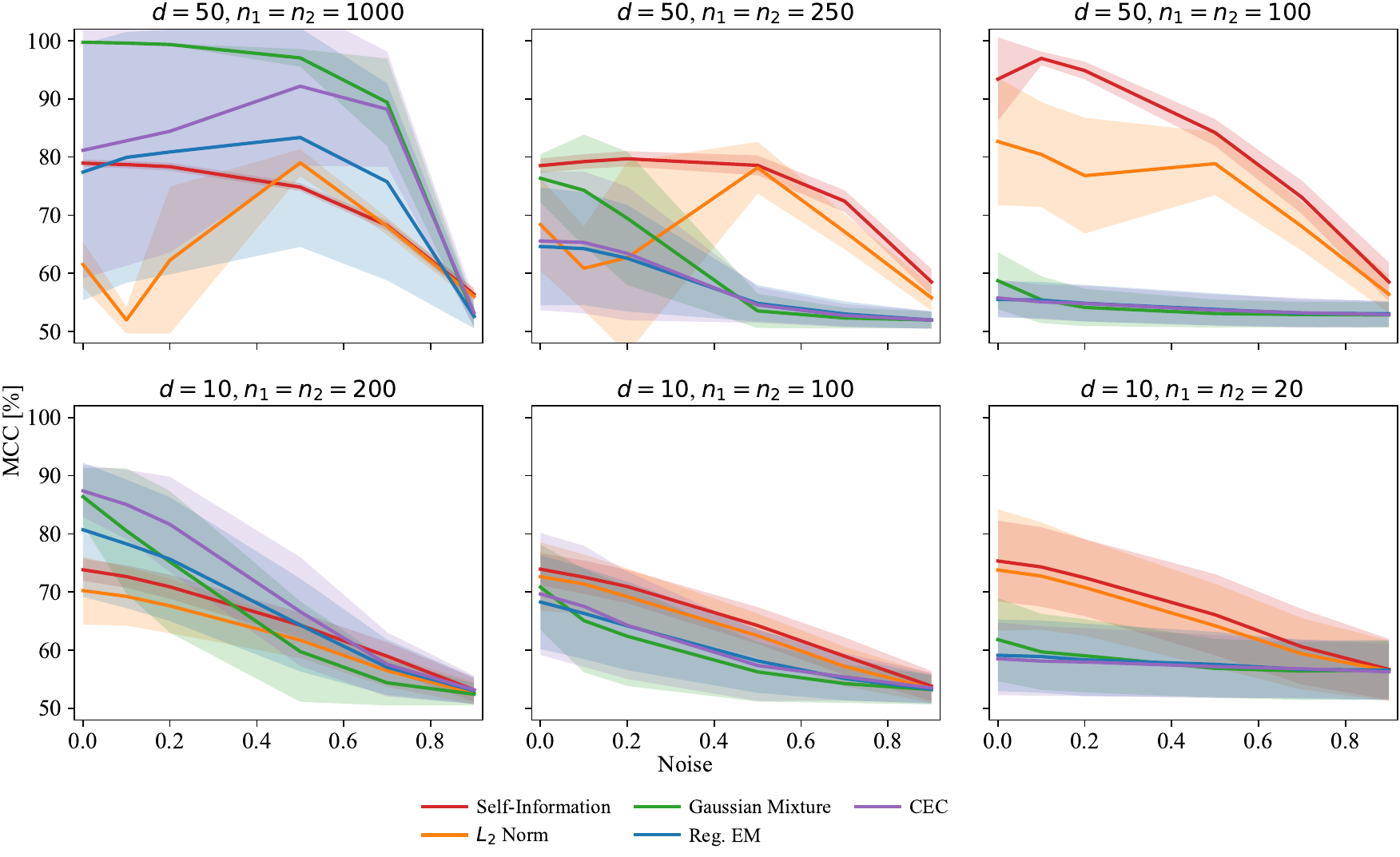}}
  \caption{Classification results of the experiment described in \S\ref{experiment_gaussian}, for a varying number of sample sizes and numbers of dimensions of multivariate Gaussian classes of varying entropy. \textbf{Upper Panels}: Varying sample sizes for $d = 50$. \textbf{Bottom Panels}: Varying sample sizes for $d = 10$. The colored areas represent one-standard-deviation intervals, derived from 100 simulations.}
     \label{fig_mcc_gaussian_dim-50}
\end{figure}

We consider a baseline dimension of $d = 50$ and evaluate performance across varying sample sizes per class. Each class is drawn from a multivariate Gaussian distribution, with both distributions centered at zero and defined by covariance matrices $\Sigma_1$ and $\Sigma_2$. The eigenvalues of these covariance matrices are set to be uniformly $\lambda_1 = 10$ and $\lambda_2 = 30$, respectively. 
To introduce noise, we modify several eigenvalues of $\Sigma_1$ to match those of $\Sigma_2$. The noise level is defined as the relative proportion of the eigenvalues in $\Sigma_1$ that are replaced by the corresponding eigenvalues in $\Sigma_2$. In Figure~\ref{fig_mcc_gaussian_dim-50}, we present the results of these experiments, showcasing the classification performance of each method under varying sample sizes, numbers of dimensions, and noise levels.

In these experiments, classification performance improves with both sample size and dimensionality, and our approach exhibits less sensitivity to the sample-to-dimension ratio than the comparison methods under the configurations tested. The two Gaussian classes share the same mean and differ primarily in their covariance spectra. Consequently, the expected difference in self-information accumulates across the eigen-directions in which their covariance scales differ, as described in Appendix~\ref{app_formalism_gaussian}. The implementation nevertheless estimates and inverts a weighted covariance matrix, applying regularization before inversion. Its performance, therefore, does not arise from avoiding covariance estimation, but from the sensitivity of the objective to covariance-scale differences. The comparable performance of the $L_2$-based objective further suggests that the radial scale contributes substantially to the separation. Figure~\ref{fig_mcc_gaussian_dim-50} thus demonstrates empirical robustness under the tested covariance-scale configurations, rather than general immunity to covariance-estimation instability.

\section{Application to Textual Data: The Priestly Source(s) in the Pentateuch} \label{results}

The composition of the Pentateuch, or Torah, has been a longstanding subject of critical scholarship, with various theories proposed to account for its linguistic diversity, stylistic inconsistencies, narrative discrepancies, and theological tensions. One of the most widely discussed models is the Documentary Hypothesis, which posits that the Torah is a composite text composed of multiple sources, each reflecting distinct historical, ideological, or scribal traditions \citep{wellhausen1885prolegomena, kuenen1886historico, holzinger1893einleitung}. Among the sources reconstructed by this framework are the Yahwist (J), the Elohist (E), the Deuteronomist (D), and the Priestly (P) source. While the precise delineation of these sources remains debated, with alternative models proposing alternative groupings or rejecting the hypothesis altogether \citep{albertz2018recent}, the existence of the P material is still widely acknowledged, owing to its distinctive style and content and relative internal consistency \citep[e.g.,][]{kuenen1886historico, holzinger1893einleitung, haran1981behind, albertz2018recent}.
The texts typically classified as P exhibit certain recurring characteristics, including highly structured formulations, genealogical lists, legal material, and a focus on ritual and cultic concerns \citep[e.g.,][]{vink1969date, albertz2018recent}. Scholars have further proposed that P itself is not monolithic but contains sub-strata, most notably distinguishing between the preceding (or succeeding) Priestly Source (P) and a later Holiness School (H) \citep[e.g.,][]{ross1997composition}, which is thought to emphasize purity laws and ethical concerns \citep[e.g.,][]{klostermann1907pentateuch, knohl2007sanctuary}. 
However, the extent to which these strata represent distinct compositional layers rather than later redactional activity remains a subject of debate, and no consensus exists on their precise historical development.
Given these complexities, our study does not assume the validity of any particular compositional model; instead, it seeks to evaluate whether quantitative, entropy-based methods can identify distinct stylistic patterns in the Pentateuch more effectively than previous computerized characterization attempts \citep{dershowitz2015computerized, yoffe2023statistical, faigenbaum2024critical}. If such patterns emerge, this may lend support to certain divisions proposed in source-critical scholarship. If they do not, it may suggest that the distinction between the various traditions, particularly between P and non-P, or P and H, is more fluid than often assumed.
We selected P as the main test case for our clustering algorithm because of its highly structured, repetitive nature, which contrasts with the more vivid, colorful style of other Pentateuchal traditions. Because our approach leverages entropy-based clustering to detect formulaic patterns, it provides an opportunity to evaluate whether the hypothesized Priestly texts exhibit distinct formulaic stylistic tendencies that set them apart from surrounding material. Additionally, if a similar distinction between P and H exists (as proposed by \citet{knohl2007sanctuary}), our method may provide an independent and complementary means of quantifying and assessing it. Notably, there is a broad literary consensus that P constitutes a stylistically distinct body of texts, often considered a literary outlier for its rigid structure, formulaic expressions, and characteristic lexical choices \citep[e.g.,][]{smith1996literary, buhler2023exploring}. Here, we consider three hypothesized partitions of the three books in the Pentateuch for evaluation:

\begin{itemize}
    \item \textbf{Genesis}: We test the distinction between genealogical lists and the surrounding narrative material. The great majority of the lists are attributed to P due to their molded expression (e.g., the Toledot formula) and emphasis on lineage continuity \citep[e.g.,][]{long1984toledotformel, plum1989genealogy}. However, a few texts belong to non-P. Hence, the focus in this case is on matters of genre and form in the light of entropy rather than source division. By isolating these passages, we assess whether their self-information properties differ from those of the surrounding material. The possibility of an internal division between P and non-P in the genealogical lists warrants further investigation.   
    
    \item \textbf{Exodus}: We analyze the division between Priestly (P) and non-Priestly (non-P) material. Priestly passages in Exodus largely consist of legal and cultic instructions, particularly concerning the construction of the Tabernacle, whereas non-Priestly sections retain a more narrative-driven style \citep{romer2009exodus, schmid2015distinguishing}. This partition has been explored \citep{yoffe2023statistical, buhler2023exploring} previously, making it a useful benchmark for evaluating our method against prior results.
    
    \item \textbf{Leviticus}: We investigate whether a distinction can be quantitatively detected between the Priestly (P) and Holiness (H) materials. While both exhibit high degrees of formulaicity, the Holiness texts are often argued to introduce distinctive stylistic and theological elements \citep{knohl2007sanctuary}. If our algorithm can recover a meaningful distinction, this may provide additional empirical support for the hypothesized P/H division.

\end{itemize}

\noindent
Each of these test cases represents a distinct type of contrast in structural predictability. In Genesis, the distinction between genealogical lists and the surrounding narrative material constitutes the most pronounced contrast in formulaicity, as the genealogies are highly repetitive and rigidly structured. In Exodus, the contrast between Priestly and non-Priestly material also exhibits a clear difference, particularly in sections related to cultic legislation and Tabernacle construction, which display strongly regularized patterns \citep{yoffe2023statistical, buhler2023exploring}. The P/H distinction in Leviticus had not previously been analyzed using computerized methods. Its inclusion here allows us to assess whether subtle stylistic differences, such as those associated with the Holiness corpus, manifest in statistically measurable differences in self-information. Together, these examples are chosen to gauge the method’s ability to quantify varying degrees of formulaicity across multiple strata of the biblical text.

\subsection{Experimental Setup} \label{results_experiment}

\subsubsection{Digital Biblical Corpus and Annotation by Experts} \label{methods_digital_corpus}

We use a digital corpus of the Masoretic variant of the Hebrew Bible in (biblical) Hebrew, which is a version of the Leningrad codex freely available by STEPBible.\footnote{\url{https://github.com/STEPBible/STEPBible-Data}} This dataset comes parsed with full morphological and semantic tags for all words, prefixes, and suffixes. In this work, we consider only the \textit{morphological} (i.e., grammatical) representation of the text, as word-based representations may introduce additional variability due to synonymy and context-dependent meanings. By focusing on morphological features, we ensure greater consistency in capturing structural patterns relevant to authorship and textual stratification.

For each hypothesized partition, we obtain expert annotations provided by biblical scholars specializing in source-critical analysis. These annotations serve as reference classifications, delineating the hypothesized textual divisions. The expert-labeled datasets allow us to evaluate the extent to which our clustering algorithm aligns with established scholarly hypotheses and assess its ability to recover traditionally proposed literary strata in an unsupervised manner.

While this study primarily employs morphological features, due to their interpretive clarity and availability in the STEPBible corpus, the method is not restricted to such representations. In settings where morphological annotation is unavailable, surface word forms can also be used. However, surface-level representations present certain challenges when evaluating formulaicity, as variation in inflection, cliticization, or syntactic context may obscure structural repetition. Morphological normalization helps reduce this variability, revealing underlying patterns of grammatical constraint. Nonetheless, there are textual domains where surface-level lexical repetition may be more meaningful.

\subsubsection{Embedding} \label{results_embedding}

We consider a cumulative-one-hot-encoded representation of the corpus $\text{D}$, such that $\text{D} \in \mathbb{R}^{n\times \chi}$, where $n$ is the number of text units, $\chi$ is the set of all unique $n$-grams in the corpus, and $\text{d}_{ij}$ represents the number of occurrences of the $j$th feature in the $i$th text unit. 

We consider a parameter space of embedding possibilities and fully explore their permutations. Specifically, each book is embedded according to a combination of several parameters across a grid of all permutations thereof: $n$, and $\ell$, spanning the following ranges: $n \in \{1,2,3,4,5 \}$ for word $n$-grams, and $\ell \in \{2, 3, 4, 6, 8, 10, 12, 14, 18, 22, 24, 26, 28 \}$, which defines the number of (overlapping) consecutive verses over which features are aggregated. This windowing approach allows the model to capture local textual dependencies by smoothing short-range fluctuations, helping to preserve stylistic continuity while preserving meaningful segment-level distinctions.
To further balance feature richness and sparsity, we restrict our analysis to the $f\in\{100, 300,500, \text{all} \}$ most frequent features in each embedding configuration. This ensures that selected features are statistically meaningful while preventing noise from rare or idiosyncratic terms that may introduce spurious clustering patterns. By systematically varying the size of the feature set, we evaluate the robustness of our method across different levels of vocabulary complexity.

This scope is motivated by the desire to exhaustively cover potential feature spaces while ensuring that the extracted features remain both meaningful and sufficiently frequent within the biblical corpus, thereby avoiding excessive sparsity due to the limited length of biblical texts \citep[e.g.,][]{antonia2014language}. Given the relatively small size of the corpus compared to modern datasets, this balance is particularly crucial in detecting meaningful linguistic patterns while maintaining statistical robustness. 

We use cumulative $n$-gram count embeddings and the Multinomial formulation (Eqs.~\ref{eq:soft_freq_multinomial}--\ref{eq:log_likelihood_multinomial}) for the biblical analyses. This representation retains repeated feature occurrences within each running window while preserving direct linguistic interpretability.
While transformer-based models have advanced text representation in many languages \citep[e.g.,][]{shmidman2022introducing}, no robust pretrained language model for biblical Hebrew is currently available. Neural embeddings would therefore require extensive domain-specific pretraining \citep[e.g.,][]{Huertas-Tato2023Oct}, making their applicability here uncertain. The cumulative-count representation instead allows the recovered clusters to be interpreted directly through their characteristic linguistic features and in relation to existing source-critical scholarship.

To further clarify the effect of parameter settings, we note that varying the $n$-gram size directly affects both sparsity and entropy in the feature space. Larger $n$-grams tend to emphasize rigid phraseology and reduce overlap across segments, making them useful for identifying formulaic repetitions. Smaller $n$-grams increase coverage but may conflate stylistically distinct phrases. Likewise, broader running windows, $\ell$, smooth over local variation and can help reveal segmental regularities, while narrower windows are more sensitive to fine-grained shifts. Feature count thresholds,$f$, balance interpretability and statistical stability by filtering out rare or idiosyncratic features. These parameters jointly determine which aspects of formulaicity are foregrounded and how they align with the text's internal segmentation.

\subsubsection{Comparison With Previous Work} \label{results_previous}

There is very limited prior work on unsupervised classification approaches for exploring hypothesized divisions of biblical texts. Most previous studies have relied on \textit{supervised} classification methods, in which textual divisions were predefined, and models were trained on manually curated feature sets \citep{radai1985genesis, dershowitz2015computerized}. These approaches often rely on cherry-picked linguistic or stylistic markers that align with existing scholarly expectations, making the claim of statistical significance somewhat circular. Since the feature selection process is influenced by prior assumptions about textual divisions, such methods do not independently verify whether distinct literary strata emerge naturally from the data.

By avoiding reliance on predefined features and instead allowing clusters to emerge based on intrinsic textual properties, we aim to assess whether computational methods can independently recover traditionally hypothesized partitions. Given the lack of prior unsupervised studies, we compare our results with previous supervised approaches to assess whether our method aligns with existing classifications or proposes alternative structures. In addition, we employ $k$-means clustering, previously used for unsupervised clustering of biblical corpora \citep[e.g.,][]{yoffe2023statistical}, as a baseline to provide a comparative reference for our information-based approach.
We also performed this analysis using GMM clustering for completeness, with results nearly identical to those of $k$-means (as observed on synthetic data; see Fig.~\ref{fig_mcc_onehot}) and are available in \ref{app_gmm}.

\subsection{Results} \label{results_results}

\begin{figure}[t!]
\centering
{\includegraphics[scale = 0.55]{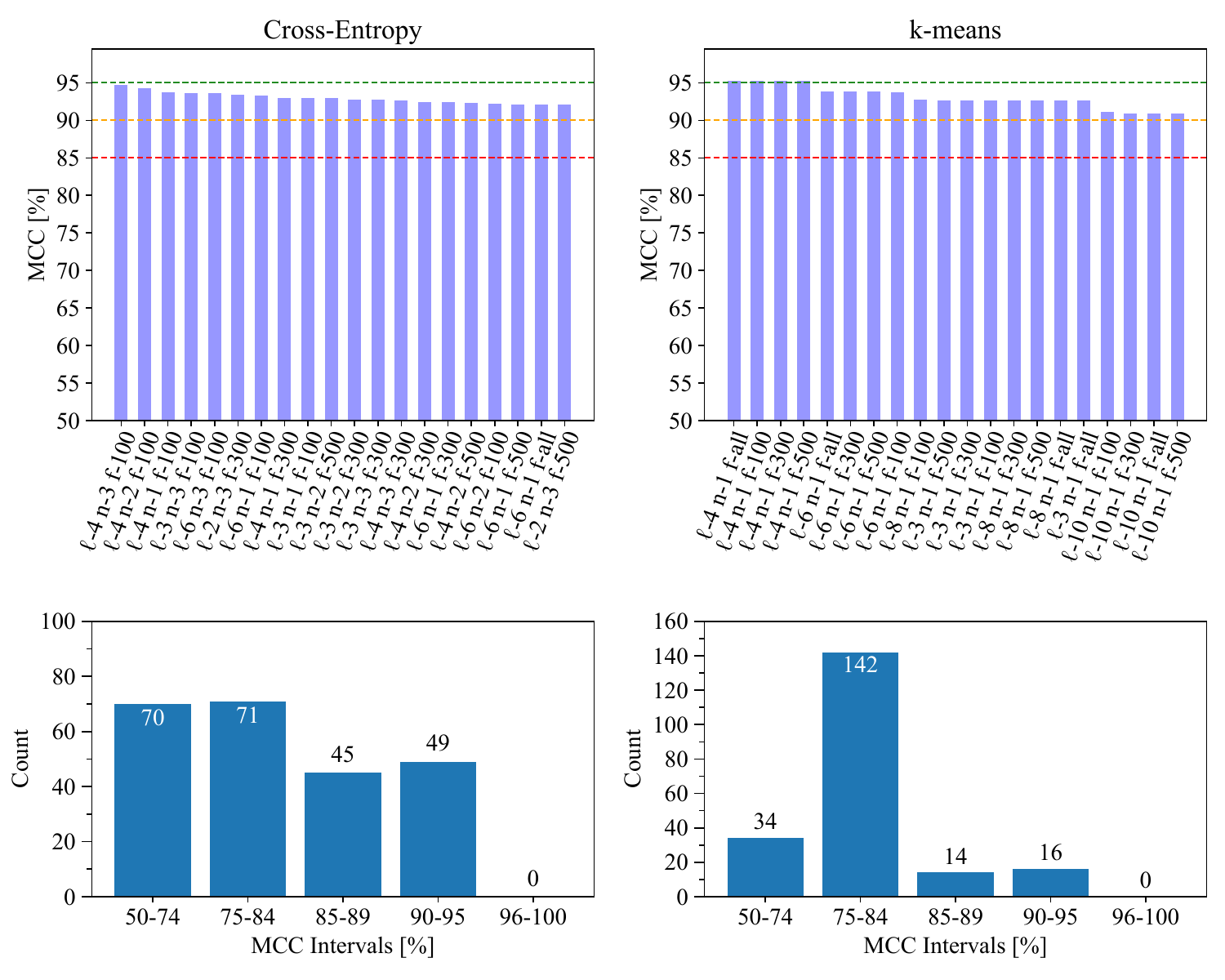}}
  \caption{Clustering results for the book of Genesis across different parameter combinations, evaluated against expert annotations distinguishing between the main textual body and genealogical lists traditionally attributed to P. Results are shown for our Multinomial cross-entropy clustering method (left) and $k$-means (right). \textbf{Top panel}: The 20 parameter combinations—defined by running-window width, $\ell$, $n$-gram size, $n$, and feature-set size, $f$—that yield the highest MCC scores. \textbf{Bottom panel}: Distribution of MCC scores across the reported parameter combinations, grouped into discrete performance intervals. Numbers displayed within or above the bars indicate the number of combinations in each interval.}
     \label{fig_genesis_results}
\end{figure}

\begin{figure}[t!]
\centering
{\includegraphics[scale = 0.55]{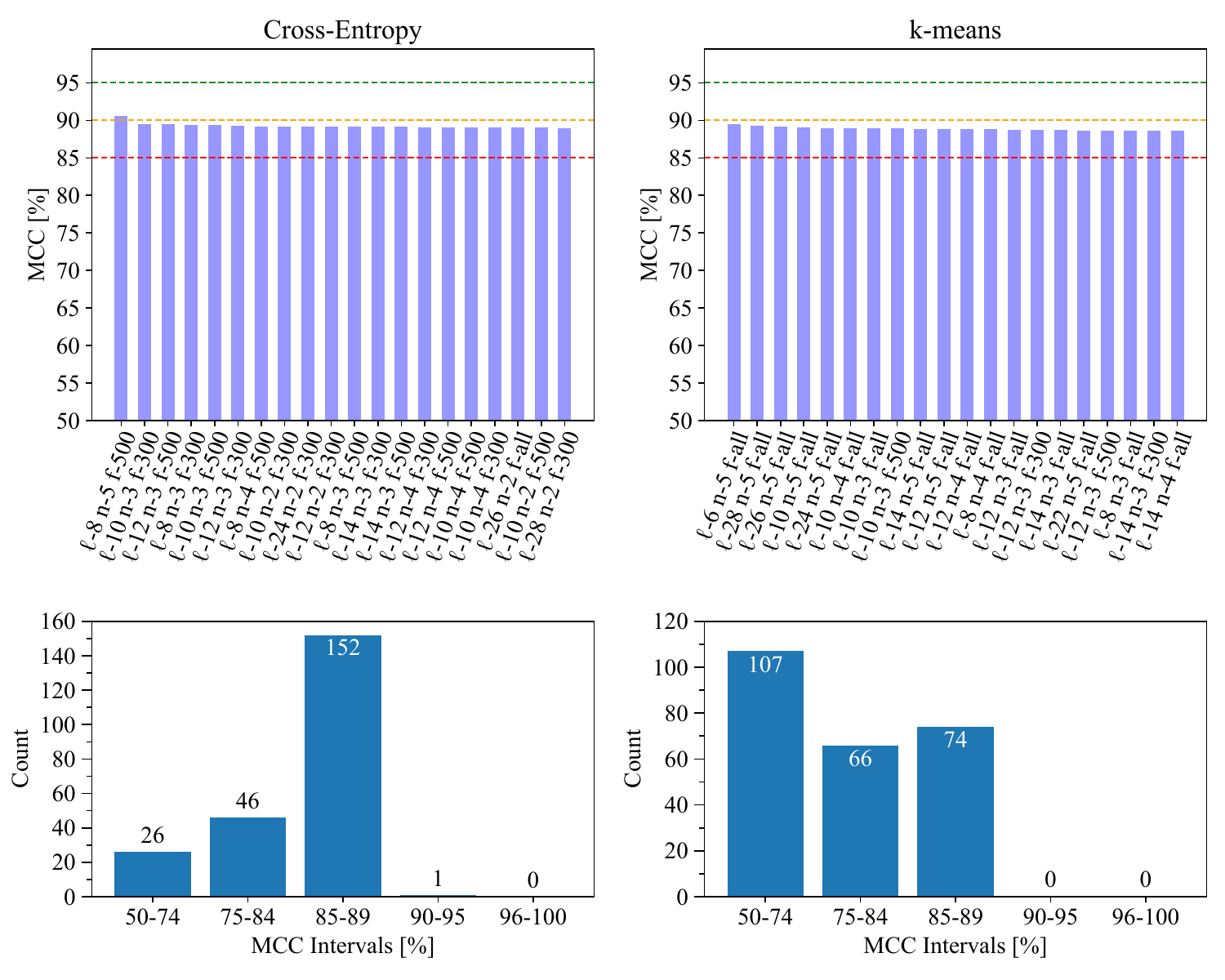}}
  \caption{Clustering results for the P/non-P partition in the book of Exodus, formatted similarly to Fig.~\ref{fig_genesis_results}.}
     \label{fig_exodus_results}
\end{figure}

\begin{figure}[t!]
\centering
{\includegraphics[scale = 0.55]{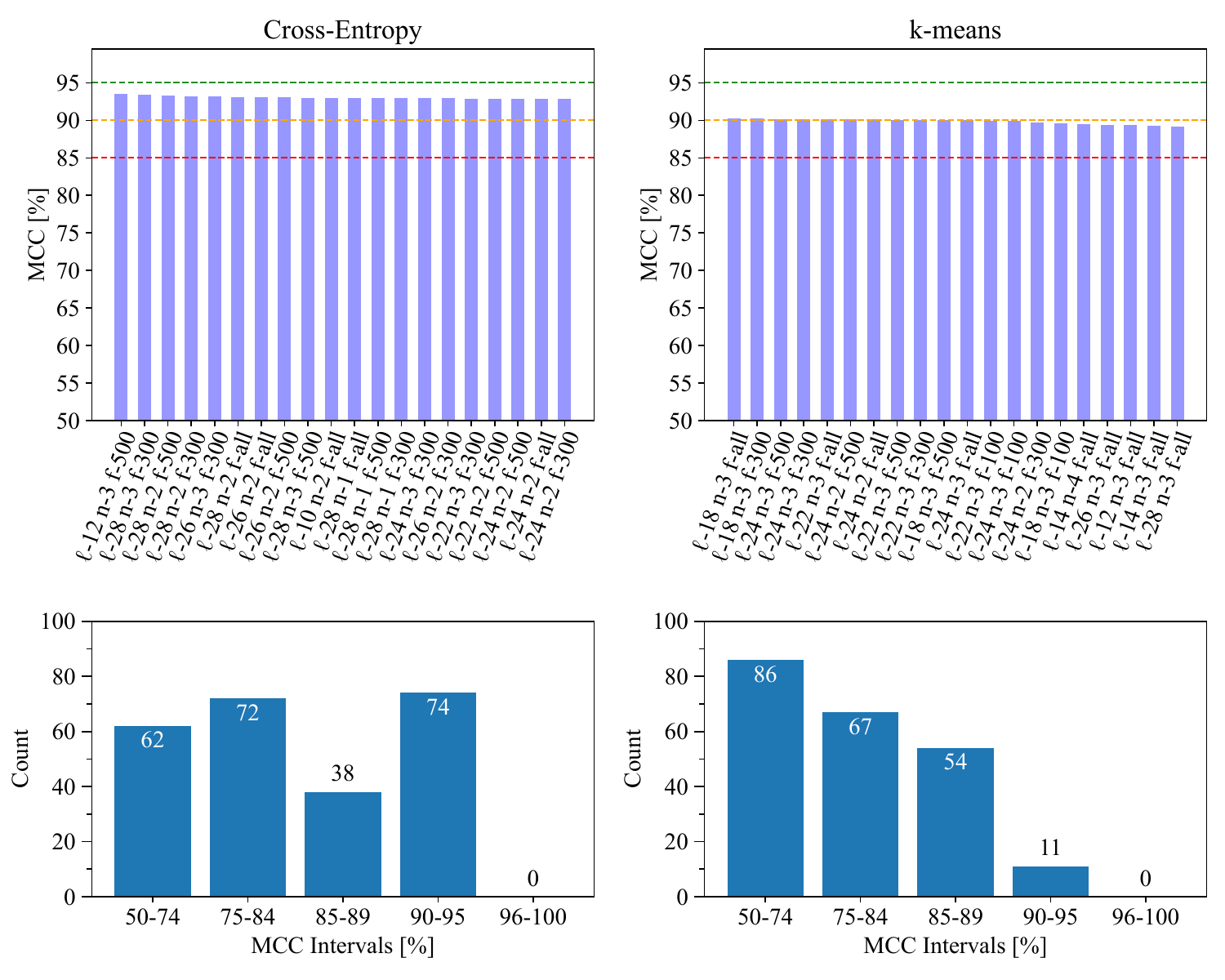}}
  \caption{Clustering results for the P/H partition in the book of Leviticus, formatted similarly to Fig.~\ref{fig_genesis_results}.}
     \label{fig_leviticus_results}
\end{figure}

The results of our clustering analysis for the hypothesized partitions of Genesis, Exodus, and Leviticus are presented in Figures~\ref{fig_genesis_results},~\ref{fig_exodus_results}, and~\ref{fig_leviticus_results}, respectively. These figures compare the performance of our information-based clustering method with $k$-means in identifying the hypothesized partitions of the text.
To assess clustering accuracy, we categorize the reported MCC scores into the intervals 50--74\%, 75--84\%, 85--89\%, 90--95\%, and 96--100\%. Because the figures report different numbers of parameter combinations for the two methods, we compare the corresponding proportions within each method. In Genesis, 94 of 235 combinations (40.0\%) obtained with our method lie in the two highest populated intervals (85--89\% and 90--95\%), compared with 30 of 206 (14.6\%) for $k$-means. In Exodus, the corresponding proportions are 153 of 225 (68.0\%) and 74 of 247 (30.0\%), respectively.

While each individual parameter combination in our grid (defined by $n$, $\ell$, and $f$) may yield results of interpretive interest, the primary objective of this study is to introduce and validate the clustering framework rather than to evaluate any specific configuration in detail. Accordingly, we assess method performance by analyzing the distribution of MCC scores across the parameter space. If a textual partition genuinely reflects differences in structural predictability, then this signal should emerge robustly across a wide range of embedding configurations, not just under fine-tuned settings.

Conversely, when only a limited subset of parameter configurations yields strong agreement with a hypothesized division, this suggests that certain levels of segmentation and feature granularity are particularly well suited to capturing certain formulaic patterns present in the data. Rather than treating this as a sign of feature selection artifacts, we interpret such behavior as evidence that distinct formulaic patterns may emerge more clearly at specific scales of representation. Some clusters, especially those linked to genre, register, or discourse framing, may only become statistically salient when, for example, the $n$-gram size and running window align with their internal structure. In this view, variation in clustering performance is not a sign of instability. Rather, it reflects both methodological sensitivity and the heterogeneous nature of formulaicity across textual layers.
This perspective is especially relevant for corpora where no predefined labels or gold standards exist. In such cases, exploring the consistency or divergence of outputs across parameter settings can help identify stable signals and locate boundaries where stylistic coherence shifts. Disagreements between parameter configurations are not necessarily failures of the method. Instead, they offer interpretive entry points into the data's structure and the scale at which formulaic patterns operate.

For the P/H partition in Leviticus, 112 of 246 combinations (45.5\%) obtained with our method lie in the two highest populated MCC intervals, compared with 65 of 218 (29.8\%) for $k$-means. Within the 90--95\% interval, the corresponding counts are 74 and 11, while 62 and 86 configurations, respectively, lie below 75\%. Thus, under the parameter grid examined here, the information-based objective recovers the expert P/H partition across a larger proportion of the reported configurations and reaches a higher maximum agreement than $k$-means. The variation across parameter settings nevertheless indicates that the recovered distinction remains sensitive to the linguistic scale at which the text is represented. In \S\ref{results_leviticus}, we examine two high-scoring configurations that foreground different forms of regularity within the P- and H-aligned material.

\subsection{Formulaic Structure and Parameter Sensitivity in the P/H Partition of Leviticus} \label{results_leviticus}

Among the three case studies, the Leviticus P/H division shows the strongest dependence on the scale at which the text is represented. In Genesis and Exodus, the contrasts between genealogical and narrative passages, and between priestly and non-priestly material, respectively, produce many strong results across the upper MCC intervals. In Leviticus, our method yields 74 configurations in the 90--95\% interval, compared with 11 for $k$-means, and 112 of 246 configurations (45.5\%) across the two highest populated intervals, compared with 65 of 218 (29.8\%). These results show that the information-based objective recovers the expert P/H partition across a larger proportion of the reported parameter combinations, while the variation among individual configurations indicates that different linguistic scales foreground different forms of regularity. The analysis that follows illustrates this dependence through two high-scoring configurations in which either the P- or H-aligned material forms the lower-self-information cluster.

\begin{figure}[t!]
\centering
{\includegraphics[scale = 0.5]{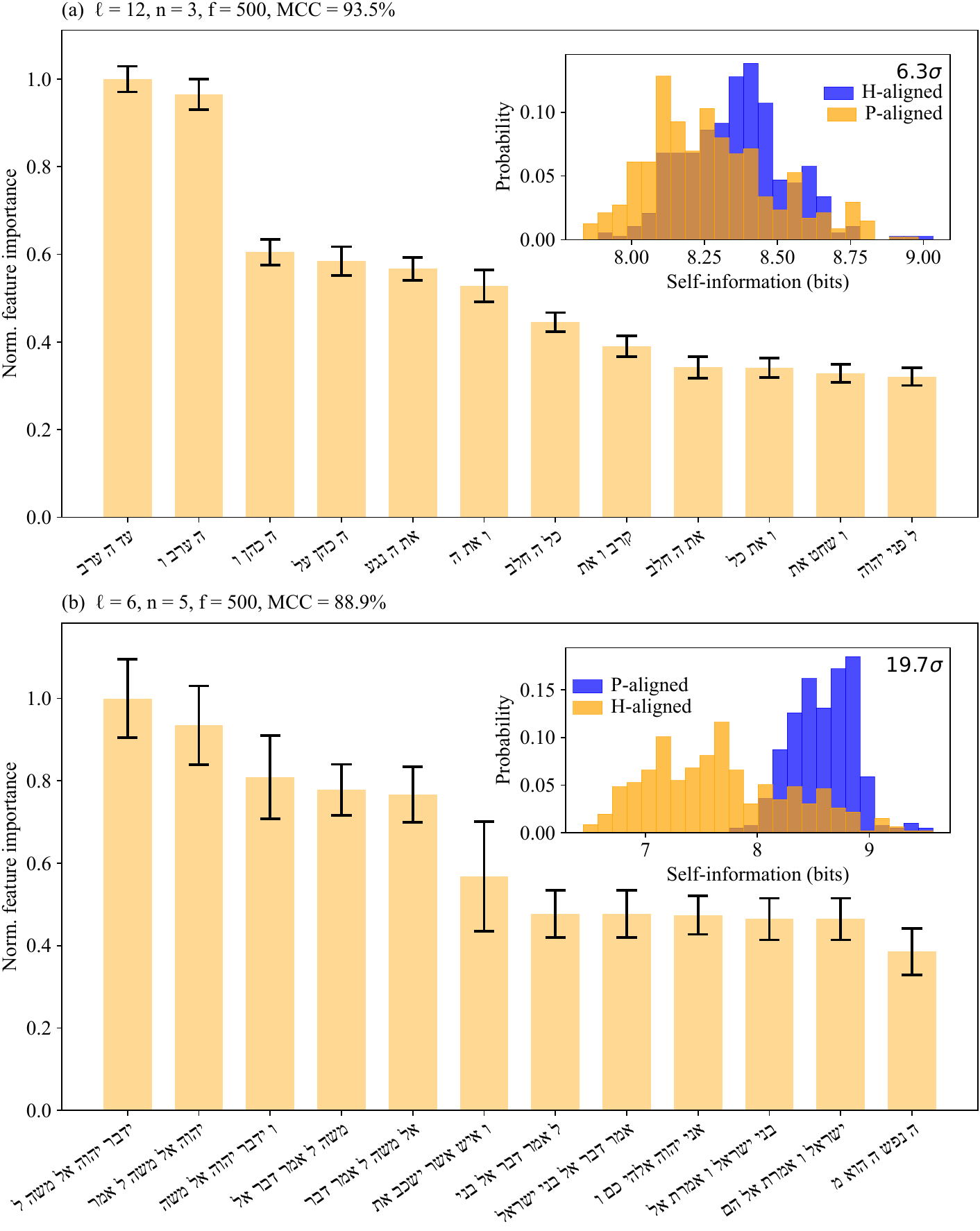}}
\caption{Distinctive morphological $n$-grams associated with the lower-self-information (formulaic) cluster in Leviticus, represented by their most frequent Hebrew surface realizations. Bars show normalized feature importance, with error bars indicating the standard deviation across 500 random half-sample estimates. \textbf{(a)} Parameter combination yielding the highest agreement with expert H/P annotations ($\ell=12$, $n=3$, $f=500$; $\mathrm{MCC}=93.5\%$); here, the P-aligned cluster has lower self-information. \textbf{(b)} Parameter combination yielding the strongest self-information separation among solutions in which the H-aligned cluster is identified as formulaic ($\ell=6$, $n=5$, $f=500$; $\mathrm{MCC}=88.9\%$). Insets show the self-information distributions of the fitted clusters. Orange denotes the lower-self-information cluster, and blue the complementary cluster; legend labels indicate their alignment with H and P. The inset annotations report the standardized separation under a permutation null ($6.3\sigma$ and $19.7\sigma$, respectively).}

     \label{fig_leviticus_features_reverse}
\end{figure}

Figure~\ref{fig_leviticus_features_reverse} shows that the alignment of the lower-self-information cluster depends on the linguistic scale of the representation: the 3-gram configuration with a 12-verse running window aligns with P, whereas the 5-gram configuration with a 6-verse window aligns with H. The configuration yielding the highest agreement with the expert P/H annotations ($n=3$, $\ell=12$, and $f=500$; $\mathrm{MCC}=93.5\%$) identifies the P-aligned component as the lower-self-information cluster. Its distinctive features comprise recurring procedural expressions rather than isolated cultic terms. These include \{\cjRL{.hlb}-\cjRL{kl h}\} (``all the fat''), which recurs in prescriptions specifying the portions removed from sacrificial animals (Lev 3:3, 3:9, and 4:8); \{\cjRL{/s.h.t 't}-\cjRL{w}\} (``and slaughter [the \ldots]''), which repeatedly introduces the slaughter of the prescribed animal (Lev 1:5, 4:4, and 4:29); and \{\cjRL{pny yhwh}-\cjRL{l}\} (``before the LORD''), which locates ritual actions within a recurring cultic setting (Lev 1:3, 1:5, 3:1, and 4:4). Likewise, \{\cjRL{`rb}-\cjRL{`d h}\} (``until evening'') repeatedly marks the duration of ritual impurity (e.g., Lev 11:27--31 and 15:17--19). Together, these action--object, locative, and temporal sequences reflect the procedural organization of the P-aligned material: its formulaicity is expressed not merely through repeated vocabulary but through compact and recurring instructions governing sacrifice and purity.

In the second case, we consider the parameter combination yielding the strongest self-information separation among solutions in which the H-aligned cluster has lower self-information ($n=5$, $\ell=6$, and $f=500$; $\mathrm{MCC}=88.9\%$; Fig.~\ref{fig_leviticus_features_reverse}b). Its distinctive features comprise overlapping morphological 5-grams whose surface realizations reconstruct recurring legislative frames. These include \{\cjRL{ydbr yhwh 'l m/sh}-\cjRL{w}\} (``And the LORD spoke to Moses[, saying]''), which marks the openings of successive legislative units (e.g., Lev 17:1, 18:1, and 19:1), and \{\cjRL{hm}-\cjRL{'ly} \cjRL{'mrt}-\cjRL{y/sr'l w} [\cjRL{dbr 'l bny}]\} (``[Speak to the] Israelites and say to them''), which introduces laws addressed to the community (e.g., Lev 18:2, 23:2, and 25:2). The resulting feature profile is more strongly descriptive of H than P. It combines recurring legislative introductions, which introduce divine laws and mark transitions between legislative units, with the theological and moral refrains characteristic of H. Their dense co-occurrence within six-verse windows reinforces divine authority and Israel’s covenantal obligations, producing a locally recurring formulaic system that distinguishes the H-aligned material from P \citep{knohl2007sanctuary}.

\section{Discussion} \label{conclusion}

In this study, we introduced an information-theoretic soft clustering framework designed to identify structured patterns within textual data. Our approach leverages self-information as a statistical indicator of structural regularities, enabling a systematic exploration of linguistic and stylistic consistency across different types of corpora. The continuous memberships serve as a relaxation of the binary clustering problem during optimization and are thresholded after convergence to obtain the recovered partition. They facilitate the search over candidate partitions but are not interpreted as posterior probabilities or as direct measures of uncertainty in textual attribution.
We validated our method on both Gaussian-simulated and categorical datasets to assess its behavior across different data structures. Under the sparse, high-dimensional categorical configurations tested, the information-based objective often recovered the planted formulaic partition more accurately than the comparison methods, although the advantage was not uniform across the full parameter range. 
The Gaussian simulations evaluated performance under controlled covariance-scale separation. Our method remained effective under several configurations in which the number of observations was small relative to the number of features. Because the continuous formulation also estimates and inverts a regularized weighted covariance matrix, this performance should be understood as an empirical property of the objective under the tested conditions, rather than as immunity to finite-sample covariance-estimation error. The comparable performance of the $L_2$-based objective further indicates that radial scale contributes substantially to the separation.

The categorical experiments, in turn, tested the framework under planted activation differences, feature dependencies, sparsity, and small sample sizes. The method remained effective across many of these configurations and often outperformed Euclidean distance-based clustering when the formulaic activation pattern was sufficiently expressed.
Applying the method to the biblical corpus, we examined whether the recovered clusters align with traditional source-critical hypotheses. Across the two highest populated MCC intervals, our method recovered 94 of 235 reported configurations in Genesis (40.0\%), 153 of 225 in Exodus (68.0\%), and 112 of 246 in Leviticus (45.5\%). The corresponding results for $k$-means were 30 of 206 (14.6\%), 74 of 247 (30.0\%), and 65 of 218 (29.8\%), respectively. These comparisons show that the information-based objective recovers the expert partitions across a larger proportion of the reported parameter combinations in all three test cases.
These findings suggest that information-theoretic clustering offers a promising avenue for unsupervised textual analysis, providing an independent measure of structural consistency without requiring the diagnostic linguistic features to be selected in advance. Its performance across the three corpora supports further application to textual datasets in which latent structures require interpretable, data-driven identification.

More broadly, our findings demonstrate that structural predictability, quantified through sample-wise self-information, captures a core dimension of textual organization that aligns with, and in some cases clarifies, traditional source-critical hypotheses. In the case of computerized source criticism of the Hebrew Bible, prior computational studies have shown that blocks such as P and non-P differ in feature distributions \citep{dershowitz2015computerized, yoffe2023statistical, faigenbaum2024critical}. Here, we show that they also differ in their degree of formulaicity. This distinction is particularly informative when made explicit through statistical modeling, as it formalizes what has often been an implicit intuition in biblical exegesis: that certain textual strata are not only lexically distinct but also stylistically more rigid, repetitive, or constrained. Crucially, our results further reveal that the expression of formulaicity depends on the feature set used; certain representations yield different interpretations of which passages are most internally consistent. This variability underscores the interpretive value of entropy-based methods: rather than producing a single fixed partition, they allow scholars to examine how different linguistic scales foreground different dimensions of textual structure, enabling a more nuanced and data-grounded exploration of compositional layers.

This makes entropy-based clustering particularly well suited for historical texts with compositional depth, where multiple stylistic logics may operate simultaneously. In such settings, soft clustering based on self-information variation provides a principled way to detect internal regularities without relying on predefined labels or explicit boundaries. Exploring the parameter space across different $n$-gram granularities, segmentation schemes, and feature types serves as an interpretive tool for probing how distinct textual regimes express constraint, formulaicity, or stylistic coherence.
For example, the distinction between Early and Late Biblical Hebrew is often expressed in systematic lexical and idiomatic shifts, making diachronic strata promising candidates for unsupervised clustering based on word-level features \citep{hurvitz1968chronological, hurvitz2014concise}. In the Deuteronomistic history, recurring thematic formulas and narrative structures may yield internally coherent segments whose predictability aligns with redactional layers \citep{peckham2019composition}. Rabbinic legal corpora, such as the Mishnah, show compositional depth through repeated conditional constructions and formulaic attributions that vary across tractates or schools \citep{neusner1988mishnah, fraade1991tradition}. In such cases, entropy-based clustering may help identify stylistic layers that are not organized by topic or subject matter, but by the underlying patterns of discourse structure and phrasing. Similarly, in oral-derived corpora like the Old Norse sagas, genealogical prologues, episodic transitions, and set-piece scenes create localized structural regularities that could be identified by entropy-guided segmentation \citep{clover1982medieval, byock1984saga}. In all these contexts, the method can be used to test source-critical hypotheses, uncover latent boundaries, or evaluate competing assumptions about where and how regularity is embedded in the textual fabric.

In future work, we will extend our framework by incorporating additional embedding techniques, such as tf-idf, neural, and transformer-based contextualized representations. These approaches will allow us to assess the model's performance across a broader range of textual features and evaluate its adaptability to modern NLP methodologies. By integrating embeddings that capture both statistical word importance and deep semantic relationships, we aim to refine the clustering process and further improve the detection of latent structural patterns in diverse textual corpora.

\section*{Acknowledgements} 

We acknowledge Prof. Israel Knohl for useful discussions and the anonymous referees for improving the quality of this manuscript.

\section*{Competing Interests}
\noindent
The authors declare none.

\section*{Data Availability Statement}
\noindent
All textual data is available online and is referenced herein. An example of our code is available at \url{https://github.com/YoffeG/cross_entropy_clustering}.

\section*{Funding Statement}
This research received no external funding.

\section*{Disclosure of use of AI tools}
The authors used AI-assisted technologies to refine the text of this manuscript.

\section*{Author Contributions}
\noindent
Conceptualization: G.Y. \\
Methodology: G.Y, B.S. \\
Software: G.Y \\
Validation: G.Y \\
Writing – Original Draft: G.Y \\
Writing – Review \& Editing: All authors

\printbibliography

@article{Stamatatos2009,
  title = {A Survey of Modern Authorship Attribution Methods},
  author = {Stamatatos, E.},
  journal = {Journal of the American Society for Information Science and Technology},
  volume = {60},
  number = {3},
  pages = {538--556},
  year = {2009},
  publisher = {Wiley Online Library}
}

@article{haran1981behind,
  title = {Behind the Scenes of History: Determining the Date of the {P}riestly Source},
  author = {Haran, M.},
  journal = {Journal of Biblical Literature},
  volume = {100},
  number = {3},
  pages = {321--333},
  year = {1981}
}

@article{hurvitz1968chronological,
  title = {The Chronological Significance of {A}ramaisms in {B}iblical {H}ebrew},
  author = {Hurvitz, A.},
  journal = {Israel Exploration Journal},
  pages = {234--240},
  year = {1968},
  publisher = {JSTOR}
}

@article{devlin2018bert,
  title={BERT: Pre-training of Deep Bidirectional Transformers for Language Understanding},
  author={Devlin, J. and Chang, M.-W. and Lee, K. and Toutanova, K.},
  journal={arXiv preprint arXiv:1810.04805},
  year={2018}
}

@book{knohl2007sanctuary,
  title={The Sanctuary of Silence: The {P}riestly {T}orah and the Holiness School},
  author={Knohl, I.},
  year={2007},
  publisher={Eisenbrauns}
}

@article{dershowitz2015computerized,
  title={Computerized Source Criticism of Biblical Texts},
  author={Dershowitz, I. and Akiva, N. and Koppel, M. and Dershowitz, N.},
  journal={Journal of Biblical Literature},
  volume={134},
  number={2},
  pages={253--271},
  year={2015},
  publisher={Society of Biblical Literature}
}

@article{shmidman2022introducing,
  title={Introducing BEREL: BERT Embeddings for {R}abbinic-Encoded Language},
  author={Shmidman, A. and Guedalia, J. and Shmidman, S. and Shmidman, C. S. and Handel, E. and Koppel, M.},
  journal={arXiv preprint arXiv:2208.01875},
  year={2022}
}

@book{holzinger1893einleitung,
  title={Einleitung in den {H}exateuch},
  author={Holzinger, H.},
  volume={1},
  year={1893},
  publisher={Mohr Siebeck}
}

@book{radai1985genesis,
  title={{G}enesis: An Authorship Study in Computer-Assisted Statistical Linguistics},
  author={Radday, Y. T. and Shore, H.},
  series={Analecta Biblica},
  volume={103},
  year={1985},
  publisher={Biblical Institution Press}
}

@inproceedings{yoffe2023statistical,
  title={A Statistical Exploration of Text Partition Into Constituents: The Case of the {P}riestly Source in the Books of {G}enesis and {E}xodus},
  author={Yoffe, G. and Bühler, A. and Dershowitz, N. and Römer, T. and Piasetzky, E. and Finkelstein, I. and Sober, B.},
  booktitle={Findings of the Association for Computational Linguistics: ACL 2023},
  year={2023},
  address={Toronto, Canada},
  publisher={Association for Computational Linguistics},
  url={https://aclanthology.org/2023.findings-acl.121},
  pages={1918--1940}
}

@article{Huertas-Tato2023Oct,
  author={Huertas-Tato, J. and Martín, A. and Camacho, D.},
  title={Understanding Writing Style in Social Media with a Supervised Contrastively Pre-Trained Transformer},
  journal={arXiv},
  year={2023},
  month={Oct},
  eprint={2310.11081},
  doi={10.48550/arXiv.2310.11081}
}

@article{antonia2014language,
  title={Language Chunking, Data Sparseness, and the Value of a Long Marker List: Explorations with Word \textit{n}-Grams and Authorial Attribution},
  author={Antonia, A. and Craig, H. and Elliott, J.},
  journal={Literary and Linguistic Computing},
  volume={29},
  number={2},
  pages={147--163},
  year={2014},
  publisher={Oxford University Press}
}

@article{buhler2023exploring,
  title={Exploring the Stylistic Uniqueness of the {P}riestly Source in {G}enesis and {E}xodus Through a Statistical/Computational Lens},
  author={Bühler, A. and Yoffe, G. and Römer, T. and Sober, B. and Finkelstein, I. and Piasetzky, E. and Dershowitz, N.},
  journal={Zeitschrift für die alttestamentliche Wissenschaft},
  volume={136},
  issue={2},
  publisher={Walter de Gruyter},
  year={2024}
}

@book{wellhausen1885prolegomena,
  title={Prolegomena to the History of {I}srael: With a Reprint of the Article {I}srael from the \textit{Encyclopaedia Britannica.}},
  author={Wellhausen, J.},
  year={1885},
  publisher={A. \& C. Black}
}

@article{albertz2018recent,
  title={The Recent Discussion on the Formation of the {P}entateuch/{H}exateuch},
  author={Albertz, R.},
  journal={Hebrew Studies},
  volume={59},
  pages={65--92},
  year={2018},
  publisher={JSTOR}
}

@incollection{vink1969date,
  title={The Date and Origin of the {P}riestly Code in the {O}ld {T}estament},
  author={Vink, J. G.},
  booktitle={The {P}riestly Code and Seven Other Studies},
  pages={1--144},
  year={1969},
  publisher={Brill}
}

@article{smith1996literary,
  title={The Literary Arrangement of the {P}riestly Redaction of {E}xodus: A Preliminary Investigation},
  author={Smith, M. S.},
  journal={The Catholic Biblical Quarterly},
  volume={58},
  number={1},
  pages={25--50},
  year={1996},
  publisher={JSTOR}
}

@article{tabor2014cross,
  title={Cross-entropy clustering},
  author={Tabor, Jacek and Spurek, Przemyslaw},
  journal={Pattern Recognition},
  volume={47},
  number={9},
  pages={3046--3059},
  year={2014},
  publisher={Elsevier}
}

@inproceedings{houdouin2023regularized,
  title={Regularized EM algorithm},
  author={Houdouin, Pierre and Ollila, Esa and Pascal, Fr{\'e}d{\'e}ric},
  booktitle={ICASSP 2023-2023 IEEE International Conference on Acoustics, Speech and Signal Processing (ICASSP)},
  pages={1--5},
  year={2023},
  organization={IEEE}
}

@article{vlassis2002greedy,
  title={A greedy EM algorithm for Gaussian mixture learning},
  author={Vlassis, Nikos and Likas, Aristidis},
  journal={Neural processing letters},
  volume={15},
  pages={77--87},
  year={2002},
  publisher={Springer}
}

@article{paquot2012formulaic,
  title={Formulaic language in learner corpora},
  author={Paquot, Magali and Granger, Sylviane},
  journal={Annual Review of Applied Linguistics},
  volume={32},
  pages={130--149},
  year={2012},
  publisher={Cambridge University Press}
}

@incollection{wood2019classifying,
  title={Classifying and identifying formulaic language},
  author={Wood, David},
  booktitle={The Routledge handbook of vocabulary studies},
  pages={30--45},
  year={2019},
  publisher={Routledge}
}

@incollection{read2008measurement,
  title={Measurement of formulaic sequences},
  author={Read, John and Nation, ISP},
  booktitle={Formulaic sequences: Acquisition, processing and use},
  pages={23--35},
  year={2008},
  publisher={John Benjamins Publishing Company}
}

@article{polak2006linguistic,
  title={Linguistic and stylistic aspects of epic formulae in ancient {S}emitic poetry and biblical narrative},
  author={Polak, Frank H},
  journal={Biblical Hebrew in Its Northwest Semitic Setting. Jerusalem: The Hebrew University Magnes Press and Winona Lake: Eisenbrauns},
  pages={285--304},
  year={2006}
}

@article{coleman2019psalmic,
  title={The Psalmic Oral Formula Revisited: A Cognitive-Performative Approach},
  author={Coleman, Stephen},
  journal={Biblical Interpretation},
  volume={27},
  number={2},
  pages={186--207},
  year={2019},
  publisher={Brill}
}

@misc{gitay1980tradition,
  title={Tradition and Interpretation: A Study of the Use and Application of Formulaic Language in the So-Called Ebed {YHWH}-Psalms},
  author={Gitay, Yehoshua},
  year={1980},
  publisher={Society of Biblical Literature}
}

@article{magoun1953oral,
  title={Oral-formulaic character of {A}nglo-{S}axon narrative poetry},
  author={Magoun Jr, Francis P},
  journal={Speculum},
  volume={28},
  number={3},
  pages={446--467},
  year={1953},
  publisher={The Mediaeval Academy of America}
}

@book{jensen1980homeric,
  title={The {H}omeric question and the oral-formulaic theory},
  author={Jensen, Minna Skafte},
  volume={20},
  year={1980},
  publisher={Museum Tusculanum Press}
}

@article{polak2017syntactic,
  title={Syntactic-Stylistic Aspects of the So-Called “Priestly” Work in the Torah},
  author={Polak, Frank H},
  journal={Le-Ma ‘an Ziony: Studies in Honor of Ziony Zevit},
  pages={345--382},
  year={2017}
}

@book{shectman2009strata,
  title={The strata of the priestly writings: contemporary debate and future directions},
  author={Shectman, Sarah and Baden, Joel S},
  volume={95},
  year={2009},
  publisher={Theologischer Verlag Z{\"u}rich}
}

@incollection{stipp201711,
  title={11 Formulaic Language and the Formation of the {B}ook of {J}eremiah},
  author={Stipp, Hermann-Josef},
  booktitle={Jeremiah’s Scriptures},
  pages={145--165},
  year={2017},
  publisher={Brill}
}

@article{warters1976formula,
  title={Formula criticism and the poetry of the {O}ld {T}estament},
  author={W{\"a}rters, William R},
  year={1976}
}

@article{shannon1948mathematical,
  title={A mathematical theory of communication},
  author={Shannon, Claude Elwood},
  journal={The Bell system technical journal},
  volume={27},
  number={3},
  pages={379--423},
  year={1948},
  publisher={Nokia Bell Labs}
}

@article{church1990word,
  title={Word association norms, mutual information, and lexicography},
  author={Church, Kenneth and Hanks, Patrick},
  journal={Computational linguistics},
  volume={16},
  number={1},
  pages={22--29},
  year={1990}
}

@inproceedings{huang2007dialect,
  title={Dialect classification on printed text using perplexity measure and conditional random fields},
  author={Huang, Rongqing and Hansen, John HL},
  booktitle={2007 IEEE International Conference on Acoustics, Speech and Signal Processing-ICASSP'07},
  volume={4},
  pages={IV--993},
  year={2007},
  organization={IEEE}
}

@article{klakow2002testing,
  title={Testing the correlation of word error rate and perplexity},
  author={Klakow, Dietrich and Peters, Jochen},
  journal={Speech Communication},
  volume={38},
  number={1-2},
  pages={19--28},
  year={2002},
  publisher={Elsevier}
}

@inproceedings{gamallo2017perplexity,
  title={A perplexity-based method for similar languages discrimination},
  author={Gamallo, Pablo and Campos, Jos{\'e} Ramom Pichel and Alegria, Inaki},
  booktitle={Proceedings of the fourth workshop on NLP for similar languages, varieties and dialects (VarDial)},
  pages={109--114},
  year={2017}
}

@inproceedings{kurzynski2023stylometry,
  title={The Stylometry of {M}aoism: Quantifying the Language of {M}ao {Z}edong},
  author={Kurzynski, Maciej},
  booktitle={Proceedings of the Joint 3rd International Conference on Natural Language Processing for Digital Humanities and 8th International Workshop on Computational Linguistics for Uralic Languages},
  pages={76--81},
  year={2023}
}

@article{altonji1996small,
  title={Small-sample bias in {GMM} estimation of covariance structures},
  author={Altonji, Joseph G and Segal, Lewis M},
  journal={Journal of Business \& Economic Statistics},
  volume={14},
  number={3},
  pages={353--366},
  year={1996},
  publisher={Taylor \& Francis}
}

@article{ashurbekova2021optimal,
  title={Optimal shrinkage for robust covariance matrix estimators in a small sample size setting},
  author={Ashurbekova, Karina and Usseglio-Carleve, Antoine and Forbes, Florence and Achard, Sophie},
  year={2021}
}

@article{bickel2008regularized,
  title={Regularized estimation of large covariance matrices},
  author={Bickel, Peter J and Levina, Elizaveta},
  year={2008}
}

@article{dilokthanakul2016deep,
  title={Deep unsupervised clustering with gaussian mixture variational autoencoders},
  author={Dilokthanakul, Nat and Mediano, Pedro AM and Garnelo, Marta and Lee, Matthew CH and Salimbeni, Hugh and Arulkumaran, Kai and Shanahan, Murray},
  journal={arXiv preprint arXiv:1611.02648},
  year={2016}
}

@article{casale2018gaussian,
  title={Gaussian process prior variational autoencoders},
  author={Casale, Francesco Paolo and Dalca, Adrian and Saglietti, Luca and Listgarten, Jennifer and Fusi, Nicolo},
  journal={Advances in neural information processing systems},
  volume={31},
  year={2018}
}

@inproceedings{yang2019deep,
  title={Deep clustering by gaussian mixture variational autoencoders with graph embedding},
  author={Yang, Linxiao and Cheung, Ngai-Man and Li, Jiaying and Fang, Jun},
  booktitle={Proceedings of the IEEE/CVF international conference on computer vision},
  pages={6440--6449},
  year={2019}
}

@article{davis2006differential,
  title={Differential entropic clustering of multivariate gaussians},
  author={Davis, Jason and Dhillon, Inderjit},
  journal={Advances in Neural Information Processing Systems},
  volume={19},
  year={2006}
}

@book{ross1997composition,
  title={The composition of the {H}oliness {C}ode ({L}ev. 17-26)},
  author={Ross, Jerome Clayton},
  year={1997},
  publisher={University of Pittsburgh}
}

@article{hartigan1979k,
  title={A k-means clustering algorithm},
  author={Hartigan, John A and Wong, Manchek A and others},
  journal={Applied statistics},
  volume={28},
  number={1},
  pages={100--108},
  year={1979},
  publisher={USA}
}

@inproceedings{ester1996density,
  title={Density-based spatial clustering of applications with noise},
  author={Ester, Martin and Kriegel, Hans-Peter and Sander, J{\"o}rg and Xu, Xiaowei},
  booktitle={Int. Conf. knowledge discovery and data mining},
  volume={240},
  number={6},
  year={1996}
}

@article{plum1989genealogy,
  title={Genealogy as theology},
  author={Plum, Karin Friis},
  journal={Scandinavian Journal of the Old Testament},
  volume={3},
  number={1},
  pages={66--92},
  year={1989},
  publisher={Taylor \& Francis}
}

@article{romer2009exodus,
  title={The {E}xodus Narrative According to the {P}riestly Document},
  author={R{\"o}mer, Thomas},
  journal={The Strata of the Priestly Writings. Contemporary Debate and Future Directions},
  pages={157--174},
  year={2009},
  publisher={TVZ}
}

@article{schmid2015distinguishing,
  title={Distinguishing the world of the {E}xodus narrative from the world of its narrators: the question of the priestly {E}xodus account in its historical setting},
  author={Schmid, Konrad},
  journal={Israel's Exodus in Transdisciplinary Perspective: Text, Archaeology, Culture, and Geoscience},
  pages={331--344},
  year={2015},
  publisher={Springer}
}

@article{faigenbaum2024critical,
  title={Critical biblical studies via word frequency analysis: unveiling text authorship},
  author={Faigenbaum-Golovin, Shira and Kipnis, Alon and B{\"u}hler, Axel and Piasetzky, Eli and R{\"o}mer, Thomas and Finkelstein, Israel},
  journal={arXiv preprint arXiv:2410.19883},
  year={2024}
}

@article{yao2015sample,
  title={Sample covariance matrices and high-dimensional data analysis},
  author={Yao, Jianfeng and Zheng, Shurong and Bai, ZD},
  journal={Cambridge UP, New York},
  year={2015}
}

@article{royer2017adaptive,
  title={Adaptive clustering through semidefinite programming},
  author={Royer, Martin},
  journal={Advances in Neural Information Processing Systems},
  volume={30},
  year={2017}
}

@inproceedings{li2020data,
  title={Data-dependent gaussian prior objective for language generation},
  author={Li, Zuchao and Wang, Rui and Chen, Kehai and Utiyama, Masso and Sumita, Eiichiro and Zhang, Zhuosheng and Zhao, Hai},
  booktitle={International Conference on Learning Representations},
  year={2020}
}

@book{kuenen1886historico,
  title={An Historico-Critical Inquiry into the Origin and Composition of the {H}exateuch ({P}entateuch and Book of {J}oshua)},
  author={Kuenen, Abraham and Wicksteed, Philip Henry},
  year={1886},
  publisher={Macmillan}
}

@book{klostermann1907pentateuch,
  title={Der {P}entateuch: Beitr{\"a}ge zu seinem Verst{\"a}ndis und seiner Entstehungsgeschicte},
  author={Klostermann, August},
  year={1907},
  publisher={G. B{\"o}hme}
}

@article{long1984toledotformel,
  title={Die toledotformel und die Literarische Struktur der priesterlichen Erweiterungsschicht im {P}entateuch},
  author={Long, Burke O and Tengstr{\"o}m, Sven and Tengstrom, Sven},
  year={1984}
}

@inproceedings{radford2021clip,
  title={Learning transferable visual models from natural language supervision},
  author={Radford, Alec and Kim, Jong Wook and Hallacy, Chris and Ramesh, Aditya and Goh, Gabriel and Agarwal, Sandhini and Sastry, Girish and Askell, Amanda and Mishkin, Pamela and Clark, Jack and others},
  booktitle={International conference on machine learning},
  pages={8748--8763},
  year={2021},
  organization={PmLR}
}

@article{giraldo2014measures,
  title={Measures of entropy from data using infinitely divisible kernels},
  author={Giraldo, Luis Gonzalo Sanchez and Rao, Murali and Principe, Jose C},
  journal={IEEE Transactions on Information Theory},
  volume={61},
  number={1},
  pages={535--548},
  year={2014},
  publisher={IEEE}
}

@book{principe2010information,
  title={Information theoretic learning: {R}enyi's entropy and kernel perspectives},
  author={Principe, Jose C},
  year={2010},
  publisher={Springer Science \& Business Media}
}

@article{hurvitz2014concise,
  title={A concise lexicon of {L}ate {B}iblical {H}ebrew},
  author={Hurvitz, A.},
  journal={Linguistic Innovations in the Writings of the Second Temple Period},
  pages={160},
  year={2014}
}

@book{peckham2019composition,
  title={The Composition of the {D}euteronomistic {H}istory},
  author={Peckham, Brian},
  volume={35},
  year={2019},
  publisher={Brill}
}

@book{neusner1988mishnah,
  title={The {M}ishnah: an introduction},
  author={Neusner, Jacob},
  year={1988},
  publisher={Bloomsbury Publishing PLC}
}

@book{fraade1991tradition,
  title={From tradition to commentary: {T}orah and its interpretation in the {M}idrash {S}ifre to Deuteronomy},
  author={Fraade, Steven D},
  volume={73},
  year={1991},
  publisher={SUNY Press}
}

@book{clover1982medieval,
  title={The medieval saga},
  author={Clover, Carol J},
  year={1982},
  publisher={Cornell University Press}
}

@article{byock1984saga,
  title={Saga form, oral prehistory, and the {I}celandic social context},
  author={Byock, Jesse L},
  journal={New Literary History},
  volume={16},
  number={1},
  pages={153--173},
  year={1984},
  publisher={JSTOR}
}
\appendix

\section{Self-Information in Multivariate Gaussian Distributions} \label{app_formalism_gaussian}

For a continuous random variable $X$ with probability density function $p(x)$, the \textit{self-information} of a realization $x$ is given by:
\begin{equation}
I(x) = -\log p(x),
\end{equation}
which quantifies the surprise or information content of an observation $x$. Lower values correspond to more probable events, and higher values correspond to less probable events.

\subsection{Self-Information of a Multivariate Gaussian Distribution}
Let $X \sim \mathcal{N}(\mu, \Sigma)$ be a $d$-dimensional multivariate Gaussian random variable with mean $\mu \in \mathbb{R}^d$ and covariance matrix $\Sigma \in \mathbb{R}^{d \times d}$, where the probability density function is given by:
\begin{equation}
p(x) = \frac{1}{(2\pi)^{d/2} |\Sigma|^{1/2}} \exp\left(-\frac{1}{2}(x-\mu)^\top \Sigma^{-1}(x-\mu)\right).
\end{equation}
Substituting this expression for $p(x)$ into the definition of self-information:
\begin{align}
I(x) &= -\log p(x) \nonumber \\
     &= -\left[ -\frac{d}{2} \log(2\pi) - \frac{1}{2} \log |\Sigma| - \frac{1}{2}(x-\mu)^\top \Sigma^{-1} (x-\mu) \right] \nonumber \\
     &= \frac{d}{2} \log (2\pi) + \frac{1}{2} \log |\Sigma| + \frac{1}{2} D_M(x),
\end{align}
where $D_M(x) = (x-\mu)^\top \Sigma^{-1} (x-\mu)$ is the Mahalanobis distance. This expression shows that the self-information of a sample from a multivariate Gaussian distribution consists of three terms:
\begin{itemize}
    \item A constant term $\frac{d}{2} \log(2\pi)$, which depends only on the dimensionality $d$.
    \item A cluster-conditioned covariance term $\frac{1}{2} \log |\Sigma|$, which depends on the determinant of $\Sigma$.
    \item A local deviation term $\frac{1}{2} D_M(x)$, which quantifies how far $x$ is from the mean $\mu$, normalized by the covariance structure.
\end{itemize}

\subsection{Resolving Power of Self-Information and Covariance Estimation}

To compare two multivariate Gaussian distributions, let $X_1 \sim \mathcal{N}(\mu_1, \Sigma_1)$ and $X_2 \sim \mathcal{N}(\mu_2, \Sigma_2)$. The difference in expected self-information between these two distributions is given by:
\begin{equation}
\Delta I = \mathbb{E}[I(x_2)] - \mathbb{E}[I(x_1)].
\end{equation}
Using the previous result for $I(x)$, we can express the expected self-information as:
\begin{align}
\mathbb{E}[I(x)] &= \frac{d}{2} \log(2\pi) + \frac{1}{2} \log |\Sigma| + \frac{1}{2} \mathbb{E}[D_M(x)].
\end{align}
Since $\mathbb{E}[D_M(x)] = d$ (the expected Mahalanobis distance for a Gaussian distribution, see \ref{app_mahalanobis}), we simplify this to:
\begin{equation}
\mathbb{E}[I(x)] = \frac{d}{2} \log(2\pi) + \frac{1}{2} \log |\Sigma| + \frac{d}{2}.
\end{equation}

Now, to compute the difference in expected self-information between the two distributions, we have:
\begin{align}
\Delta I &= \left[ \frac{d}{2} \log(2\pi) + \frac{1}{2} \log |\Sigma_2| + \frac{d}{2} \right] 
- \left[ \frac{d}{2} \log(2\pi) + \frac{1}{2} \log |\Sigma_1| + \frac{d}{2} \right] \nonumber \\
&= \frac{1}{2} \log \frac{|\Sigma_2|}{|\Sigma_1|}.
\end{align}

Since the determinant of a matrix $\Sigma_k$ is the product of its eigenvalues, $|\Sigma_k| = \prod_{i=1}^d \lambda_{k,i}$, where $\lambda_{k,i}$ are the eigenvalues of $\Sigma_k$, we can rewrite the difference as:
\begin{equation}
\Delta I = \frac{1}{2} \sum_{i=1}^d \ln \frac{\lambda_{2,i}}{\lambda_{1,i}}.
\end{equation}

If the eigenvalue ratios $\lambda_{2,i} / \lambda_{1,i}$ are approximately constant across dimensions, say $\lambda_{2,i} / \lambda_{1,i} \approx C$, then:
\begin{equation}
\Delta I \approx \frac{d}{2} \ln C.
\end{equation}

Thus, the resolving power of self-information increases with the dimensionality $d$, allowing for a better distinction between two distributions.

\subsection{Variance of Self-Information and Finite-Sample Averaging}

For fixed and known parameters $\mu$ and $\Sigma$, the squared Mahalanobis distance of a Gaussian observation follows a chi-squared distribution:
\begin{equation}
D_M(X) \sim \chi_d^2.
\end{equation}
Because $\operatorname{Var}(D_M(X))=2d$, and the only random term in the self-information is $\frac{1}{2}D_M(X)$, the variance of the self-information of an individual observation is
\begin{equation}
\operatorname{Var}(I(X))=\frac{1}{4}\operatorname{Var}(D_M(X))=\frac{d}{2}.
\end{equation}
This variance does not decrease with sample size and does not depend on the overall scale of the covariance eigenvalues.

The dependence on sample size arises when considering the mean self-information of $n$ independent observations,
\begin{equation}
\overline{I}_n=\frac{1}{n}\sum_{r=1}^{n}I(X_r).
\end{equation}
For fixed Gaussian parameters,
\begin{equation}
\operatorname{Var}(\overline{I}_n)
=\frac{\operatorname{Var}(I(X))}{n}
=\frac{d}{2n}.
\end{equation}
Thus, increasing the sample size reduces the uncertainty in the estimated mean self-information, rather than the intrinsic observation-to-observation variance of $I(X)$. When $\mu$ and $\Sigma$ are themselves estimated from finite data, covariance-estimation uncertainty introduces additional effects that are not represented by this expression.

\subsection{Impact on Resolving Power and Signal-to-Noise Ratio (SNR)}

We define the resolving power for distinguishing two distributions from their sample-mean self-information values as
\begin{equation}
\text{Resolving Power}
=\frac{|\Delta I|}
{\sqrt{\operatorname{Var}(\overline{I}_{1,n})+
\operatorname{Var}(\overline{I}_{2,n})}}.
\end{equation}
For two independent samples of equal size $n$ and dimensionality $d$, this becomes
\begin{equation}
\text{Resolving Power}
=|\Delta I|\sqrt{\frac{n}{d}}.
\end{equation}
If the covariance-eigenvalue ratios are approximately constant, $\lambda_{2,i}/\lambda_{1,i}\approx C$, then $\Delta I\approx \frac{d}{2}\ln C$, and therefore
\begin{equation}
\text{Resolving Power}
\approx \frac{|\ln C|}{2}\sqrt{nd}.
\end{equation}

\begin{figure}
\centering
{\includegraphics[scale = 0.52]{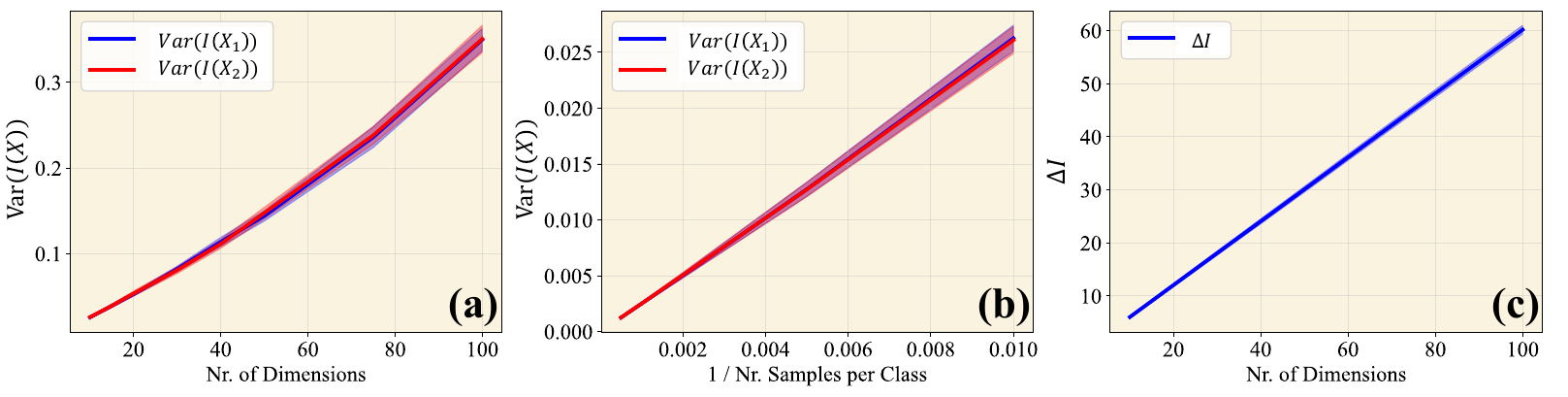}}
  \caption{Dependence of sample-mean self-information on sample size and dimensionality. \textbf{(a)} Linear dependence of $\operatorname{Var}(\overline{I}_n)$ on dimension at fixed $n$. \textbf{(b)} Linear dependence of $\operatorname{Var}(\overline{I}_n)$ on $1/n$ at fixed $d$. \textbf{(c)} Linear dependence of $\Delta I$ on dimension for an approximately constant covariance-eigenvalue ratio.}
     \label{fig_gaussian_var_theory}
\end{figure}

\noindent
Figure~\ref{fig_gaussian_var_theory} illustrates the corresponding dependence of the sample-mean variance on $d$ and $n$, and the linear dependence of $\Delta I$ on $d$.

\section{Expected Mahalanobis Distance} \label{app_mahalanobis}

Let $X \sim \mathcal{N}(\mu, \Sigma)$ be a $d$-dimensional Gaussian random variable with mean vector $\mu$ and covariance matrix $\Sigma$. The Mahalanobis distance of a realization $x$ is defined as:
\[
D_M(x) = (x - \mu)^\top \Sigma^{-1} (x - \mu).
\]
The quantity $D_M(x)$ measures how far the vector $x$ is from the mean $\mu$, normalized by the covariance matrix $\Sigma$.

Since $x$ is drawn from a Gaussian distribution, we can express $D_M(x)$ as the sum of squared standard normal variables:
\[
D_M(x) = \sum_{i=1}^d Z_i^2,
\]
where $Z_i \sim \mathcal{N}(0, 1)$ are independent standard normal random variables, each representing the standardized components of the vector $x - \mu$ under the covariance structure defined by $\Sigma$.

The expectation of the squared standard normal variables is:
\[
\mathbb{E}[Z_i^2] = 1 \quad \text{for each} \quad i = 1, 2, \dots, d.
\]
Therefore, the expected Mahalanobis distance is the sum of the expectations for each component:
\[
\mathbb{E}[D_M(x)] = \mathbb{E}\left[ \sum_{i=1}^d Z_i^2 \right] = \sum_{i=1}^d \mathbb{E}[Z_i^2] = d.
\]

Thus, the expected value of the Mahalanobis distance for a sample from a multivariate Gaussian distribution is $d$, which depends solely on the dimensionality of the distribution:
\[
\mathbb{E}[D_M(x)] = d.
\]

\section{Distinguishing Entropy-Based and Feature Distribution-Based Clusters} \label{app_feat_norm}

While the previous categorical benchmark (\S\ref{experiment_onehot}) demonstrated the effectiveness of our approach in distinguishing formulaic from non-formulaic structures in sparse, high-dimensional settings, an additional challenge arises when two clusters exhibit identical entropy yet differ in their feature distributions. In real-world textual corpora, stylistic or formulaic patterns are not always characterized by large entropy contrasts but may instead be defined by shifts in the distribution of certain linguistic elements. If our method is to serve as a robust tool for identifying latent structures in textual data, it must be capable of distinguishing between two cases: (1) when formulaic patterns manifest as an overall reduction in entropy, and (2) when distinct compositional layers exist despite maintaining equivalent entropy levels.

\subsection{Experimental Setup}

To investigate this, we construct a controlled dataset in which two classes share identical overall entropy but differ in their feature distributions. We then introduce a formulaic subset by modifying the activation probabilities of a selected fraction of features. This experiment tests whether the self-information-based clustering framework correctly resolves compositional differences when entropy alone does not provide a discriminative signal. By evaluating the model’s classification performance in both the entropy-based and feature-distribution-based scenarios, we assess the extent to which our approach captures deeper structural regularities beyond simple variance constraints.

\paragraph{Feature Probability Distributions.} 

Given a binary feature space $X \in \{0,1\}^{n \times d}$ with $n$ samples and $d$ dimensions, we construct two distinct classes, $A$ and $B$, while ensuring that their overall entropy remains identical.

Initially, all dimensions are assigned a base activation probability $p$. To create distinct but entropy-balanced distributions, half of the dimensions are randomly selected and modified as follows:

\begin{itemize}
    \item For the selected dimensions, class $A$ retains the original activation probability $p$, while class $B$ receives a reduced probability of $0.1p$.
    \item The remaining half of the dimensions follow the inverse assignment: class $B$ retains the original activation probability $p$, while class $A$ receives $0.1p$.
\end{itemize}

Formally, let $S_{\text{high}} \subset \{1, \dots, d\}$, $|S_{\text{high}}| = d/2$ be a randomly selected subset of dimensions. The activation probabilities for each feature in each class are then:

\[
p_A(j) = 
\begin{cases} 
p, & j \in S_{\text{high}} \\
0.1p, & j \notin S_{\text{high}}
\end{cases}, 
\quad 
p_B(j) = 
\begin{cases} 
0.1p, & j \in S_{\text{high}} \\
p, & j \notin S_{\text{high}}
\end{cases}.
\]

Samples for each class are then drawn from independent Bernoulli distributions:

\[
X_A(j) \sim \text{Bern}(p_A(j)), \quad X_B(j) \sim \text{Bern}(p_B(j)).
\]

Since both classes contain the same number of dimensions at probability $p$ and $0.1p$, their total entropy remains identical:

\[
H(X_A) \approx H(X_B).
\]

\paragraph{Formulaic Cluster Construction.} 

To introduce a formulaic cluster, a subset of each class ($n/2$ samples) is modified by increasing the activation probability of a selected fraction $d_{\mathrm{form}}$ of dimensions. These dimensions, referred to as \textit{formulaic dimensions}, are chosen randomly from the total feature set.

The activation probabilities of these formulaic dimensions are modified as follows:

\[
p_A'(j) = 
\begin{cases}
\min\{1,p_A(j) + p_{\mathrm{form}}\}, & j \in S_{\text{form}} \\
p_A(j), & \text{otherwise}
\end{cases}, 
\quad
p_B'(j) = 
\begin{cases}
\min\{1,p_B(j) + p_{\mathrm{form}}\}, & j \in S_{\text{form}} \\
p_B(j), & \text{otherwise}
\end{cases},
\]

where $S_{\text{form}} \subset \{1, \dots, d\}$ is a randomly selected subset of dimensions with size $d_{\mathrm{form}}d$, and $p_{\mathrm{form}}$ is the activation-probability increment applied to those dimensions. The minimum ensures that all activation probabilities remain bounded by one.

\paragraph{Labeling.} 

Each sample is labeled according to two independent classification axes:

\begin{itemize}
    \item \textbf{Feature distribution label:} $Y_{\text{dist}} \in \{0,1\}$ where 0 corresponds to class $A$ and 1 to class $B$.
    \item \textbf{Formulaicity label:} $Y_{\text{form}} \in \{0,1\}$ where 0 corresponds to original samples and 1 to the formulaic subset.
\end{itemize}

This formulation allows evaluation of whether the clustering framework differentiates formulaic structure from feature-distribution-based separation.

\subsection{Experimental Results}

To assess the capacity of the information-theoretic clustering framework to distinguish between formulaic and feature-distribution-based structures, we analyze clustering performance as a function of key dataset parameters. Specifically, we vary the baseline feature-activation probability ($p_{\mathrm{feature}}$), the activation-probability increment applied to formulaic dimensions ($p_{\mathrm{form}}$), and the fraction of dimensions designated as formulaic ($d_{\mathrm{form}}$). 

\begin{figure}
    \centering
    \includegraphics[scale = 0.3]{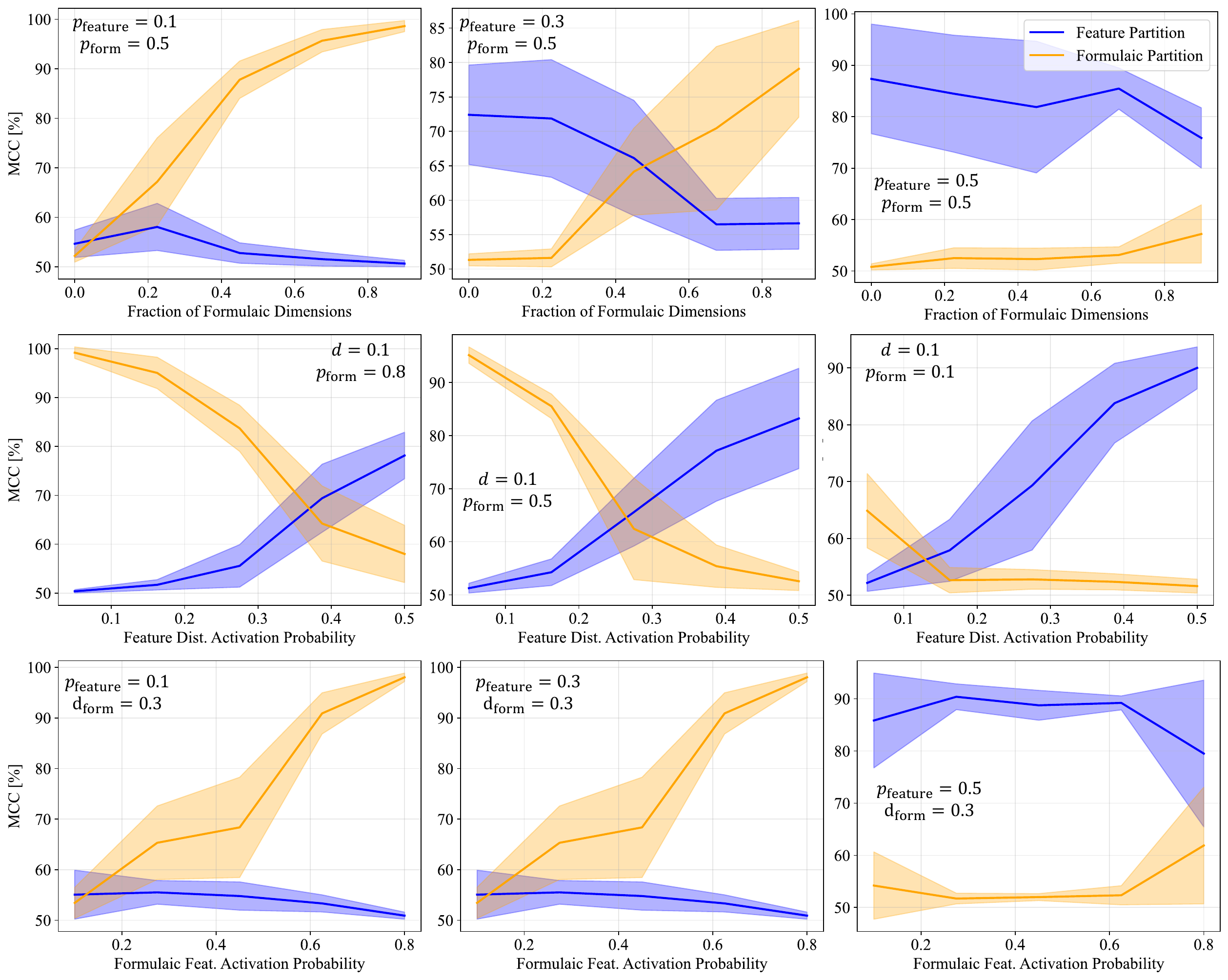}
    \caption{
    Clustering performance, quantified by the normalized label-invariant MCC, as a function of key dataset parameters. Orange shows agreement between the recovered partition and the formulaicity labels $Y_{\mathrm{form}}$, whereas blue shows agreement with the feature-distribution labels $Y_{\mathrm{dist}}$. Both curves evaluate the same clustering output against the two independently planted partitions.
    \textbf{Upper panel:} MCC versus the fraction of formulaic dimensions $d_{\mathrm{form}}$, with fixed activation increment $p_{\mathrm{form}}=0.5$ and varying $p_{\mathrm{feature}}\in\{0.1,0.3,0.5\}$.
    \textbf{Middle panel:} MCC versus the baseline feature-activation probability $p_{\mathrm{feature}}$, with fixed $d_{\mathrm{form}}=0.1$ and varying $p_{\mathrm{form}}\in\{0.8,0.5,0.1\}$.
    \textbf{Lower panel:} MCC versus the formulaic-feature activation increment $p_{\mathrm{form}}$, with fixed $d_{\mathrm{form}}=0.3$ and varying $p_{\mathrm{feature}}\in\{0.1,0.3,0.5\}$.
    }
    \label{fig:ent_features_plot}
\end{figure}

Figure~\ref{fig:ent_features_plot} compares the agreement of the same recovered partition with two independently planted structures: the formulaicity labels $Y_{\mathrm{form}}$ and the feature-distribution labels $Y_{\mathrm{dist}}$. The comparison therefore evaluates which structural signal dominates the recovered partition under each parameter regime; it does not compare two separate clustering methods. This analysis remains a heuristic demonstration rather than a quantitative predictive framework, and does not establish a universal boundary between the two regimes.

As the formulaic signal is strengthened through $d_{\mathrm{form}}$ or $p_{\mathrm{form}}$, the recovered partition generally becomes more strongly aligned with $Y_{\mathrm{form}}$. Conversely, when the global feature-distribution contrast dominates, agreement with $Y_{\mathrm{dist}}$ is higher. The same clustering output can therefore align with either planted partition, depending on the relative statistical strength of the partitions. These results illustrate the interaction between the two signals under the simulated conditions, but do not define general criteria for predicting which will dominate in other datasets.

\section{GMM Clustering Results} \label{app_gmm}

\begin{figure}[t!]
\centering
\rotatebox{90}{\includegraphics[scale = 0.35]{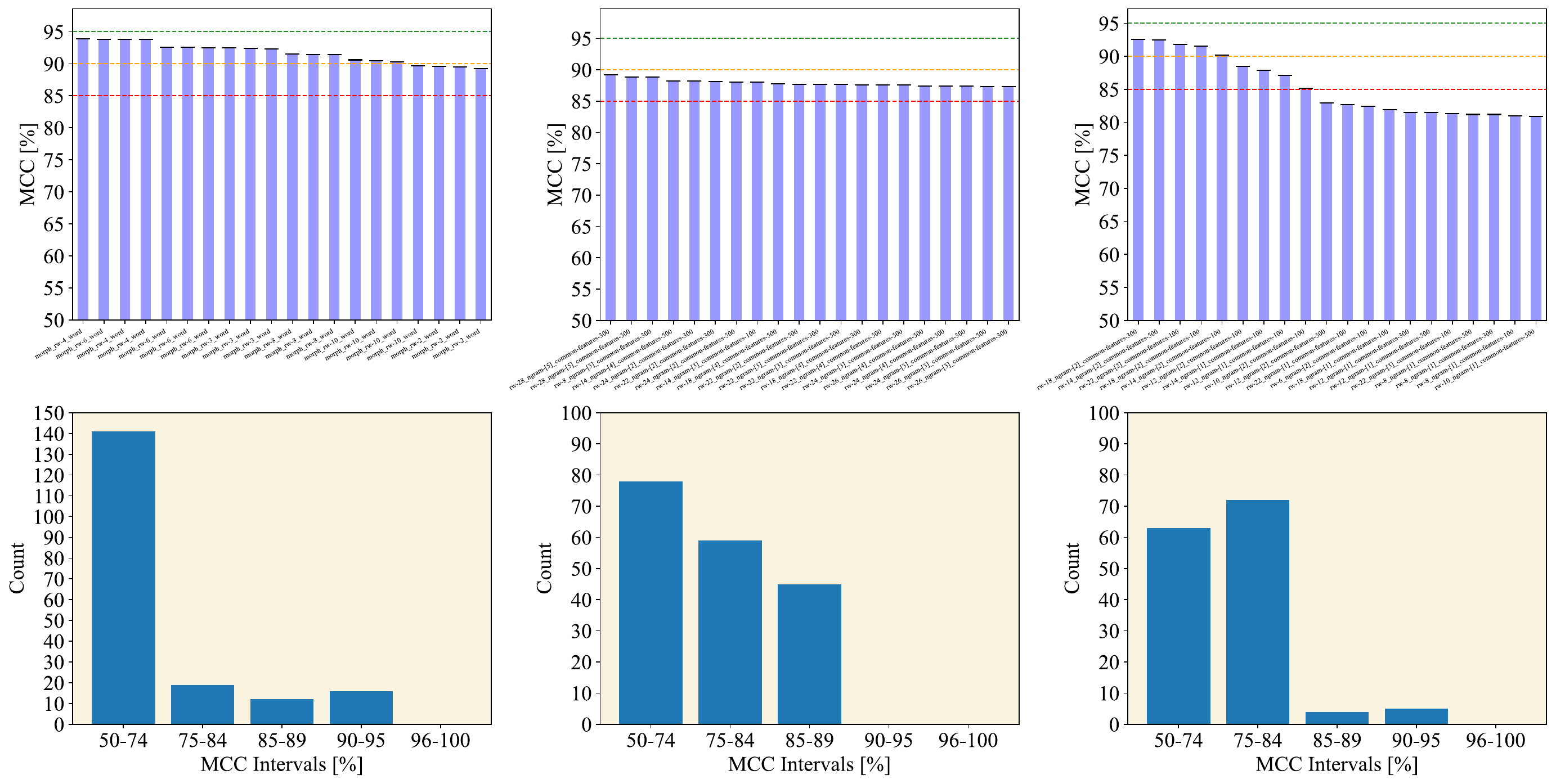}}
  \caption{Clustering results for the hypothesized partitions of the books of Genesis, Exodus, and Leviticus (left to right, respectively), using GMM clustering, similar to Fig.~\ref{fig_genesis_results}.}
     \label{fig_gmm}
\end{figure}

In Figure~\ref{fig_gmm}, we illustrate the classification results using the Gaussian Mixture Model (GMM) algorithm as applied to the three biblical texts, as discussed in \S\ref{results}.

\end{document}